\pdfoutput=1

\documentclass[11pt]{article}

\usepackage{EACL2023}

\usepackage{times}
\usepackage{latexsym}

\usepackage[T1]{fontenc}

\usepackage[utf8]{inputenc}
\usepackage{booktabs}

\usepackage{microtype}

\usepackage{inconsolata}

\usepackage{graphicx}
\usepackage{caption}
\usepackage{subcaption}
\usepackage{amsfonts}

\title{Paparazzi: A Deep Dive into the Capabilities of Language and Vision Models for Grounding Viewpoint Descriptions}

\author{
  Henrik Voigt \\
  Affiliation / Address line 1 \\
  \texttt{henrik.voigt@uni-jena.de} 
  \\\And
  Jan Hombeck \\
  Affiliation / Address line 1 \\
  \texttt{jan.hombeck@uni-jena.de} 
  \\\And
  Monique Meuschke \\
  Affiliation / Address line 1 \\
  \texttt{meuschke@isg.cs.uni-magdeburg.de} 
  \\\And
  Kai Lawonn \\
  Affiliation / Address line 1 \\
  \texttt{kai.lawonn@uni-jena.de} 
  \\\And
  Sina Zarrieß \\
  Affiliation / Address line 1 \\
  \texttt{sina.zarriess@uni-bielefeld.de} \\\And
 }
 
 \author{
  Henrik Voigt\textsuperscript{1},
  Jan Hombeck\textsuperscript{1},
  Monique Meuschke\textsuperscript{3}, 
  Kai Lawonn\textsuperscript{1} and \textbf{Sina Zarrieß\textsuperscript{2}}  \\
  \textsuperscript{1}University of Jena  
  \textsuperscript{2}University of Bielefeld  \textsuperscript{3}University of Magdeburg \\
  \textsuperscript{1}\texttt{first.last@uni-jena.de} \\ \textsuperscript{2}\texttt{first.last@uni-bielefeld.de} \\
  \textsuperscript{3}\texttt{last@isg.cs.uni-magdeburg.de}
}

\begin{document}
\maketitle
\begin{abstract}
Existing language and vision models achieve impressive performance in image-text understanding. Yet, it is an open question to what extent they can be used for language understanding in 3D environments and whether they implicitly acquire 3D object knowledge, e.g. about different views of an object.
In this paper, we investigate whether a state-of-the-art language and vision model, CLIP, is able to ground perspective descriptions of a 3D object and identify canonical views of common objects based on text queries.
We present an evaluation framework that uses a circling camera around a 3D object to generate images from different viewpoints and evaluate them in terms of their similarity to natural language descriptions.
We find that a pre-trained CLIP model performs poorly on most canonical views and that fine-tuning using hard negative sampling and random contrasting yields good results even under conditions with little available training data.
\end{abstract}


\section{Introduction}
\label{sec:introduction}
\begin{figure}
 \centering 
 \includegraphics[width=\linewidth]{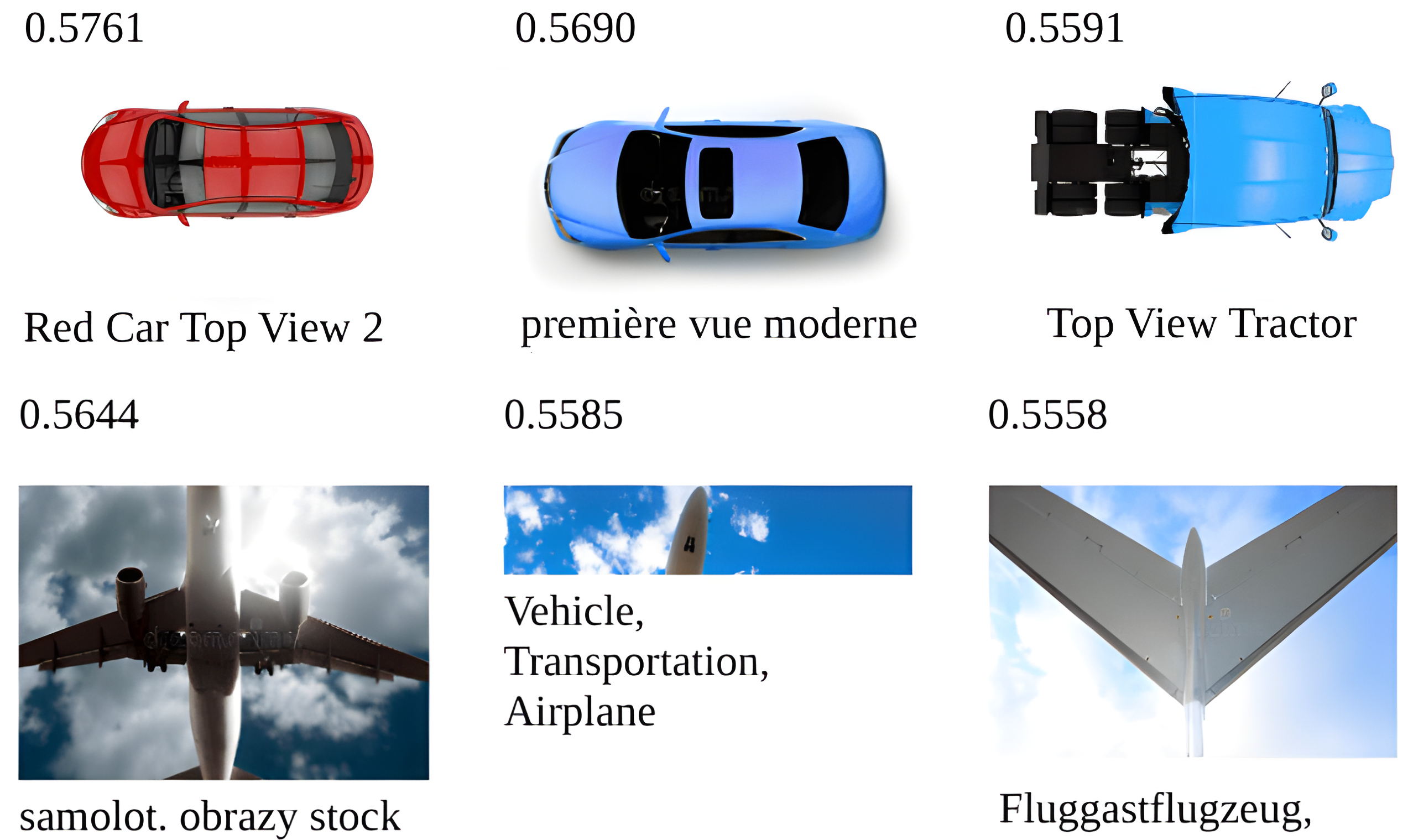}
\caption{Top-3 retrieval results for \textit{car/airplane from the bottom} using CLIP on the LAION-5B dataset.\footnotemark}
 \label{fig:laion_search_example}
\end{figure}

Recent advancements in pre-training large-scale language and vision (L\&V) models, such as CLIP~\cite{radford2021learning}, have led to exceptional performance on benchmarks and leaderboards in 2D image-text retrieval~\cite{shen2021much, fang2021clip2video, baldrati2022effective}. 
However, the image-text data in these benchmarks have specific properties and biases~\cite{thomason2022language} that may limit the language grounding capabilities of existing L\&V models and their robustness in real-word scenarios~\cite{khandelwal2022simple, gadre2022clip}. A fundamental bias in existing L\&V data comes from the fact that images generally show single, human-centric views of \textit{different objects}.
This raises a simple but intriguing question:
to what extent can a model acquire knowledge about the concept of viewpoints and identify \textit{different} views on the \textit{same object}?
%
Figure \ref{fig:laion_search_example} illustrates this challenge, showing the top-3 images retrieved by CLIP for two basic viewpoint descriptions, \textit{car/airplane from the bottom}, in the LAION-5B~\cite{schuhmann2021laion}  data set:  the \textit{airplane} images mostly correspond to the correct view, but none of the \textit{car} images shows a bottom view. 
It suggests that the model does not generalize the meaning of viewpoint descriptions across different objects,\footnote{When searching the LAION-5B dataset via image embeddings of cars from the bottom, dozens of relevant results can be provided, which shows that these views exist in the data.} and may fail to acquire visual-linguistic knowledge that would be needed in more realistic 3D scenarios, such as when instructing a drone to take a picture of an object from a specific viewpoint~\cite{thomason2020vision, fan2022aerial}.
This opens the door for a systematic examination of the capabilities of L\&V models for grounding viewpoint descriptions, delving into the question of why, despite their excellent zero-shot capabilities, a model like CLIP struggles when it comes to representing perspectives of the same object. 

%
\footnotetext{\url{https://rom1504.github.io/clip-retrieval/}}

In this paper, we investigate whether language understanding in pre-trained L\&V models generalizes to simple text-viewpoint descriptions of common objects.
We propose a new task -- text-viewpoint retrieval -- and a framework for analyzing and scaling image-text models with 3D data. We implement a \textbf{\texttt{Paparazzi}} agent that circles a  spherical camera around a 3D object, samples images, and scores pairs of image-viewpoint descriptions using a pre-trained image-text matching model. In this framework, we evaluate and analyze whether CLIP, as a representative image-text-matching model with excellent zero-shot capabilities, systematically retrieves images of views of 3D shapes, regardless of potential reporting biases in 2D L\&V data sets.

To successfully interpret viewpoint descriptions like \textit{car from the bottom}, models need to connect concepts in natural language to visual representations and basic knowledge of object geometry.
To investigate this, our approach is deliberately simple: we use 3D shapes from five categories of common objects in ShapeNet that have visually distinct canonical views \textit{(front, back, left, right, top, bottom)}.
Based on Goldberg polyhedrons~\cite{goldberg1937class}, that divide a sphere into hexagonal shapes, we analyze whether CLIP provides an adequate embedding for the viewpoint space around an object.
Our analysis suggests that basic viewpoint understanding is indeed a systematic gap in the pre-trained CLIP model, as it achieves very poor performance in scoring view-description pairs and even retrieves nonsensical, non-human-centric views.
Furthermore, we find that this problem is not fixed by standard fine-tuning.
Thus, we propose a procedure for fine-tuning CLIP that extends the contrastive learning approach to viewpoints and descriptions generated from 3D visualizations.
We find that a small amount of training data and extended fine-tuning is successful in scaling CLIP to basic viewpoint understanding in 3D.
\section{Related Work}
\label{sec:related_work}


\paragraph{Vision, View, and Language.}
To date, research on grounding language in vision focuses on connecting language to visual representations of 2D human-centric views of scenes and objects based on, e.g., large image-caption data sets~\cite{thomee2016yfcc100m, schuhmann2021laion}. 
Retrieval models in L\&V usually rank a fixed set of images showing single views of different objects and scenes given a textual query or vice versa \cite{li2020unicoder, li2020oscar, baldrati2022effective}. Common understanding models process pairs of texts or questions and single-view images and predict labels for them, typical generation models process single-view images and generate descriptions for them~\cite{mokady2021clipcap, yu2022coca}. In this paper, we propose a new L\&V retrieval task where the model needs to search for a specific view, represented as an image, of a 3D object given a textual query. In our task, the space of possible view-images is not restricted to a human-centric view.

\paragraph{Language Grounding in 3D.}
\citet{achlioptas2019shapeglot} present pioneering work in this area, with a referring expression data set designed for learning the language of shape for \textit{chair} objects in ShapeNet, the most well-known resource for 3D object models~\cite{chang2015shapenet}. They build a neural resolution model that predicts which chair is referred to by a given shape description. Their encoder combines an autoencoder for point clouds of 3D shapes and a pre-trained image encoder for a single view of the object. As \citet{achlioptas2019shapeglot} collected descriptions of the 3D objects in a static environment with a fixed camera perspective, their approach does not account for dynamic viewpoints in 3D.  \citet{thomason2022language} present a larger data set for expressions referring to ShapeNet objects and build a model that relies on image-text matching via the CLIP architecture, similar to ours. Their model takes images of eight fixed viewpoints of the object as input and integrates a component that estimates the viewing angle of an image. They evaluate on resolution accuracy and do not explicitly test viewpoint understanding in the CLIP model. In contrast to these existing works, the input to our model does not specify a fixed set of camera positions, and the output is an explicit, specific viewpoint of an object represented as an image. 



\paragraph{Camera Position Estimation.}
\label{sec:optimal_camera_position_estimation}
Viewpoint selection in a 3D environment is a well-known problem in other areas \cite{kamada1988simple,roberts1998viewpoint,arbel1999viewpoint,vazquez2001viewpoint,plemenos2006viewpoint, podolak2006planar,muhler2007viewpoint}. Work in photogrammetry investigates camera position estimation minimizing the error in 3D measurements and reconstruction~\cite{olague2002optimal}.  
Systems in visualization aim to find an optimized viewpoint with the least possible occlusion and maximum information content for polygonal data~\cite{vazquez2001viewpoint, neugebauer2013amnivis, meuschke2017automatic}, volumetric data~\cite{bordoloi2005view} and vector fields~\cite{lee2011view, tao2012unified}.
A key challenge in these areas is the definition of what actually constitutes a good viewpoint \cite{bonaventura2018survey}.
Most algorithms aim to find a viewpoint that is of high interest to the user~\cite{leifman2016surface, neugebauer2013amnivis}, but do not yet incorporate textual descriptions of viewpoints. 
In addition, most of these algorithms require expensive annotated mesh representations of 3D objects.
L\&V models pre-trained on raw image-text data constitute an extremely promising direction here, provided that they are capable of viewpoint understanding.



\section{Text-Viewpoint Retrieval Task}
\label{sec:text_viewpoint_retrieval_task}
We study viewpoint understanding from descriptions  and describe a framework for text-viewpoint retrieval. 
We present a task definition, the set-up of the 3D environment and the camera, and our approach to evaluation and analysis.


\subsection{Task Definition}
\label{sec:task}
We define the input of our viewpoint retrieval task to consist of a 3D scene with a single object $O$, a search query describing a viewpoint $q$, and an orbital camera $\mathit{C}$ circling the object.
The camera returns single views of the object $v$ that are represented as RGB images. 
The retrieval model's task is to find a viewpoint $v$ that matches the query $q$.
In this work, we implement retrieval via a scoring function $S$ that passes pairs of images $v$ (taken by the camera) and queries $q$ to a pre-trained text-image matching model. 
The parameterization of the orbiting camera $\mathit{C}$ determines the space of possible viewpoints $\mathit{V}$ that the retrieval model has to search. The parameter setup we used in this work is explained in detail below. 

This setting leverages the well-understood image-text matching in 2D for language grounding in 3D. Our retrieval model does not have a symbolic or explicit representation of the object's geometry but can perceive it by taking images from various perspectives. This framework is independent of different types of 3D data and only requires an engine that renders images of 3D environments.


\subsection{Camera Set-up}
\label{sec:camera}
\begin{figure}[t]
 \centering 
 \includegraphics[width=\linewidth]{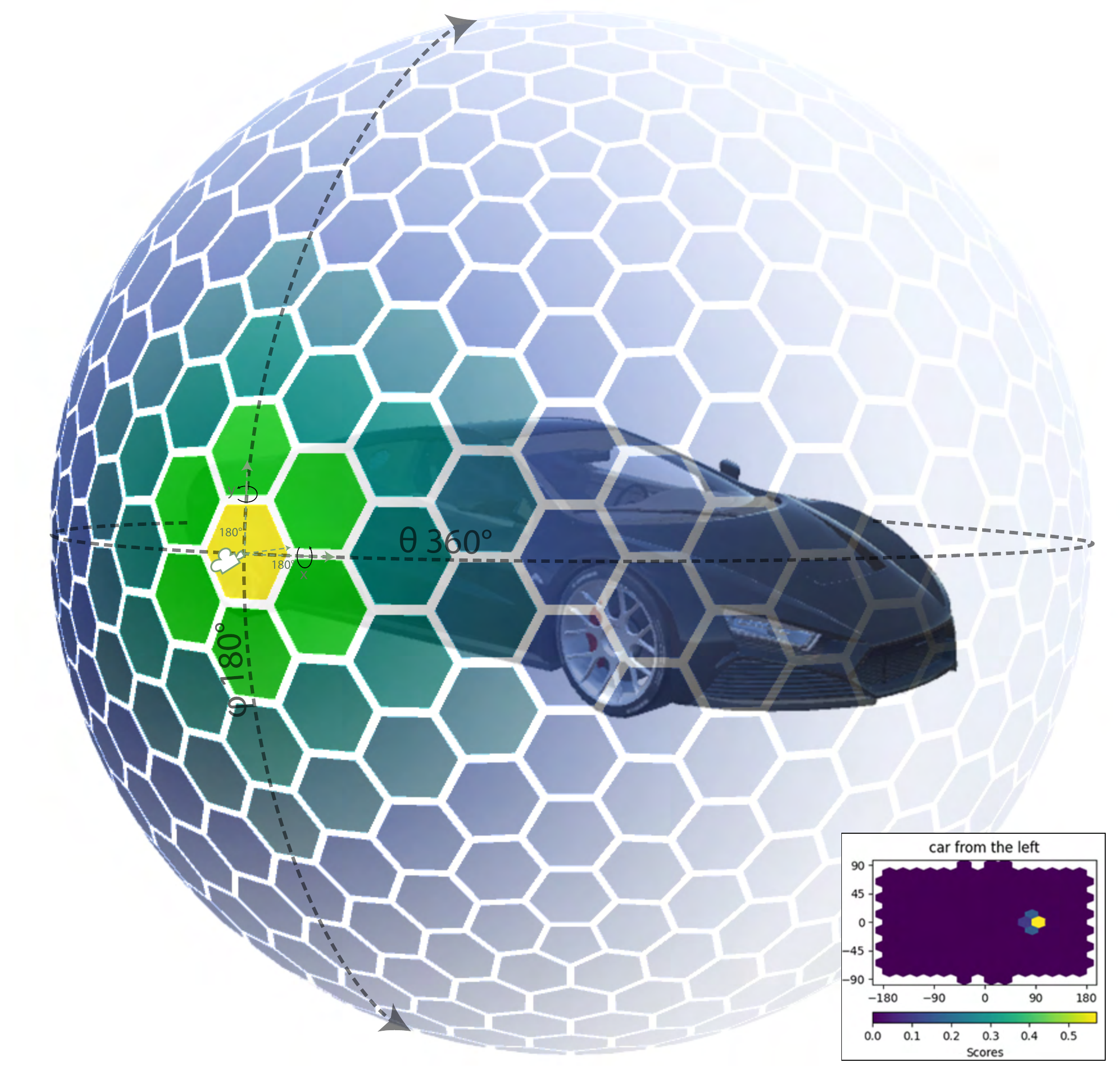}
 \caption{The camera setup: the viewing angles $\theta$ and $\varphi$ describe the azimuthal and polar angle of the camera on the orbital sphere. The parameters x and y describe the camera's orientation at the given location.}
 \label{fig:circling_camera_setup}
\end{figure}

For the purpose of this study, we restrict the viewpoint space $V$ to views that contain the object of interest.
We use a spherical camera system where the center of the object defines its center, as shown in Figure~\ref{fig:circling_camera_setup}.
The camera in orbit can be navigated around the desired object using polar coordinates. 
The position of the camera towards the object is defined by $(r,\theta,\varphi)$ for the radial distance, the azimuthal angle, and the polar angle. 
The center of the object is defined by the center of its bounding box. 
The camera's local x and y axes are used to adjust the camera's viewing angles. 
Rotation around the local z-axis of the camera is disabled in this work, as the results would be the same, only with a rotated output image. In summary, the exact camera position and rotation along the sphere can be described by five parameters: $(r,\theta,\varphi,x,y)$.

To create equidistant sample points for camera positions along the sphere, we use a Goldberg polyhedron \cite{goldberg1937class}. It divides a sphere into mostly hexagonal shapes, including a small finite number of pentagons, and creates a nearly equidistant sample space (see Figure~\ref{fig:circling_camera_setup}). The centers of the hexagons give us a discrete number of sample points, which reduce the possible configurations of our camera setup to a finite number. The hexagon centers can be approached for different radii $\mathit{r}$. The polyhedron used in this work initially yields 1002 sample points per radius. This discretization of the sample space is fine enough to allow benchmarking and analysis of viewpoint retrieval models.


The object $\mathit{O}$ lies at the origin of the Cartesian space $(0,0,0)$, which is also the center of the surrounding hypersphere. 
The radius $\mathit{r}$ is clipped relatively to the size of the object. We estimate the extent of the object based on its bounding box. We determine the extent of the bounding box based on the minimum $r_{min}$ and maximum radius $r_{max}$ of the surrounding orbital spheres. 
In our experiments, we set  $r_{min}$ to two times the edge length of the bounding box and $r_{max}$ to ten times the edge length of the bounding box.



\subsection{Evaluation and Analysis}
\label{sec:metrics}

\paragraph{Common Objects and Canonical Views.}
\label{sec:shapes}
To systematically evaluate language-view understanding in CLIP, we limit the set of viewpoint descriptions $\mathit{Q}$ in our experiments to the six \textbf{canonical views} \textit{front, back, right, left, top, bottom}  defined by~\citet{chang2015shapenet}.
We choose 3D models of common object categories in ShapeNet~\citep{chang2015shapenet}. From the available 55 categories, we selected five categories where all canonical views are visually distinct: \textit{cars, airplanes, motorbikes, mugs} and \textit{benches}.\footnote{Many object categories like \textit{bottle, ball, table, etc.} do not have this property. For instance, the \textit{front} and \textit{back} views of a bottle are not or much less distinct than the \textit{front} and \textit{back} views of a car.}
As ShapeNet provides an aligned representation of all 3D models, 
these restrictions yield a fully controllable experimental setup where training and test data with pairs of queries and views can be generated automatically. 
The experimental setup is general enough to be transferable to arbitrary object domains and various forms of textual viewpoint descriptions. 

\paragraph{Viewpoint Quality Evaluation.}
To assess the quality of text-viewpoint retrieval, we use the KL divergence~\cite{kullback1951information} of a model's scoring function against a gold standard scoring distribution as well as the classical retrieval metrics \textit{precision@k} and \textit{retrieval@k}.
We use KL divergence in addition since retrieval metrics only reflect performance on gold standard viewpoints and do not allow us to infer the global performance needed to find out why models fail on certain queries, as discussed in Section~\ref{sec:experiments}.
We define the gold standard score distribution with respect to a particular viewpoint as a discrete normal distribution around the gold standard viewpoint, which is the mean of the distribution. The three polygonal rings around the mean are assigned the normalized score value at one, two, or three times the standard deviation of the normal distribution. The scores for all these viewpoints sum to 1. The scores for all other viewpoints around the sphere are set to zero. The setup is illustrated in Figure~\ref{fig:circling_camera_setup}. 
To visually analyze the goodness of a scoring function over a sphere, we unfold the polyhedron and upsample it, as shown in the small map at the bottom right of Figure~\ref{fig:circling_camera_setup}. In this way, we can visualize the difference between the gold standard and the predicted score distribution for an object.

\paragraph{Search Performance Evaluation.}
When searching a 3D scene, there are many possible viewpoints to consider. A scoring function that works well on a subset of pre-selected viewpoints may yield a good result in retrieval metrics, but in practical usage, it may lead the search algorithm to an unexpected or nonsensical viewpoint. Therefore, to evaluate the performance of a model, we need to consider not only how well it performs on the gold standard viewpoint images, but also how well it can guide a search algorithm to find the right viewpoint in the scene. 
We compare the performance of different search algorithms under different configurations of the scoring function to \textbf{understand the impact of the shape of the scoring function on search performance}.
We compute search performance as follows: a search is considered successfully completed if the found viewpoint is within a certain radius of the respective gold standard viewpoint. We define the radius discretely based on the hexagonal rings around a gold standard viewpoint on the Goldberg polyhedron. In our experiments, we consider a search to be solved if a viewpoint is found within the first two rings around the gold standard viewpoint (see Figure~\ref{fig:circling_camera_setup}). We compare performance in terms of the number $c$ of calls to the scoring function required by the search algorithm to solve the search problem described above. We restrict the search length to a maximum number $c_{max}$ of 300 viewpoints to visit. To obtain a robust comparison, we run the procedure $n$ times at randomly selected starting positions on the hypersphere around the object. In our experiments, we set $n$ to ten. Then, the number of calls $\frac{c}{n}$ is averaged.


\section{Model}
\label{sec:model_architecture}

\subsection{Scoring Function}
The heart of our retrieval model is a function $S$ that outputs matching scores for pairs of images and queries $(v,q)$. 
Pre-trained L\&V models like CLIP~\cite{radford2021learning} embed $(v,q)$ pairs into a common subspace, resulting in latent vector representations $\mathbf{z_v}$ and $\mathbf{z_q}$, e.g., of size $512$ in the original CLIP.  
The output of the scoring function $S$ is the cosine similarity of the latent representations of the viewpoint image and the search query:
\begin{equation}\resizebox{.99\hsize}{!}{$
  {S}(v,q) = cos(z_v,z_q) = \frac{\mathbf{z_v} \cdot \mathbf{z_q}}{|\mathbf{z_v}||\mathbf{z_q}|}
  =\frac{\sum_{i=1}^{N} z_{v_{i}} z_{q_{i}}}{\sqrt{\sum_{i=1}^{N} z_{v_{i}}^{2}} \sqrt{\sum_{i=1}^{N} z_{q_{i}}^{2}}}
  $}
\end{equation}
To evaluate a given viewpoint with respect to a query, both are encoded into their latent representations $\mathbf{z_v}$ and $\mathbf{z_q}$, and the cosine similarity of their latent representations is used as a \textbf{score} for how well the view matches the query.

\subsection{Objective Functions} 
To achieve high similarity between associated texts and images, \citet{radford2021learning} apply a contrastive learning paradigm. In a training batch of N image-text pairs, a cosine similarity score is computed for each possible text-image combination. This leads to ${N \times N}$ scores over which a cross-entropy loss is calculated across the rows and columns. For corresponding text-image pairs, the maximum class score is expected, while for all other pairs, a minimum score is targeted. 

We extend this contrastive learning paradigm for fine-tuning CLIP with 3D data by minimizing the combination of three different loss objectives: a) for negative examples, b) for random examples, and c) for hard negative examples. 

\paragraph{Cross-Entropy Loss on Negative Examples} is calculated and summed for both queries $\mathbf{q}$ and viewpoints $\mathbf{v}$ as $\mathit{L_{\mathbf{v},\mathbf{q}}}$. The parameter $\tau$ is a learnable parameter for scaling the logits:
\begin{equation}
\mathit{L_{\mathbf{v},\mathbf{q}}}=-\frac{1}{N} \sum_{i=1}^{N} \log \frac{\exp \left(\cos\left(z_{v_{i}}, z_{q_{i}}\right) / \tau\right)}{\sum_{j=1}^{N} \exp \left(\cos\left(z_{v_{i}}, z_{q_{j}}\right) / \tau\right)}
\end{equation}

\paragraph{Cross-Entropy Loss on Random Examples} is denoted as $\mathit{L_{r}}$ and computed between annotated viewpoints and randomly generated viewpoints of the 3D scene. $\mathit{L_{r}}$ is computed exactly as in equation (2), but the contrastive examples are random images from the scene in this case. 

\paragraph{Cross-Entropy Loss on Hard Negative Examples}
referenced as $\mathit{L_{h}}$ uses images that have a different annotation but appear to be similar in latent space~\cite{li2021align}.  ~\citet{robinson2020contrastive} present a sampling method that rescales the loss of negative examples based on their similarity to the gold standard sample. Following this, the loss $\mathit{L_{h}}$ is calculated as the weighted contrastive loss between the positive samples $x^{+}$ and the hard negative samples $x^{-}$ drawn from the modified negative sampling distribution $q$:
\begin{equation}
\resizebox{.999\hsize}{!}{
$\mathit{L_{h}}= \mathbb{E}_{x^{+}\sim p_{x}^{+}} \left [-\log \frac{e^{f(x)^{T} f\left(x^{+}\right)}}{e^{f(x)^{T} f\left(x^{+}\right)}+G \mathbb{E}_{x^{-} \sim q}\left[e^{f(x)^{T} f\left(x^{-}\right)}\right]}\right]$
}
\end{equation}
In notation, $p^{+}$ is the marginal distribution of positive examples in the overall distribution of samples $p$. $q$ is the distribution of negative samples. $x$ is a single sample, $x^{+}$ and $x^{-}$ are the respective positive and negative samples. $f$ is a similarity measure, in our case it is cosine similarity. 
$\mathit{G}$ is a weighting parameter that can be used to adjust the hardness of the negative sampling. 

The total loss is parameterized as the weighted sum of the three objectives:
\begin{equation}
    L_{total} = \alpha \mathit{L_{\mathbf{v},\mathbf{q}}} + \beta \mathit{L_{r}} + \gamma \mathit{L_{h}}
\end{equation}
The ablations resulting from the different combinations presented above are evaluated in Section~\ref{sec:analysis}.
The parameters $\alpha$, $\beta$, and $\gamma$ are chosen based on the respective experiment.

\subsection{Search Algorithms}
\label{sec:appendix_b}
At inference time, our retrieval model requires a search algorithm $\mathit{A}$, a function that optimizes the output of the scoring function $\mathit{S}$ given the space of viewpoints $V$ and a query $q$. We compare the performance of two search algorithms. \textbf{Greedy search} starts with a grid-based approach on the Goldberg polyhedron and tries to find the optimum by moving greedily in the direction of the neighboring region with the highest score in each iteration. \textbf{Bayesian search} samples positions on the hypersphere based on incrementally obtained function values, attempting to sample with higher probability in regions that contain optima~\cite{mockus1994application}. See appendix~\ref{sec:appendix_a} for implementation details.

\section{Experiments}
\label{sec:experiments}
\subsection{Experimental Setup}
\label{sec:problem_setup}
\paragraph{Training.} For each of the six canonical view query types and five object categories, we generate 1,000 training images in a Unity scene on randomly selected objects from the ShapeNet training set.
This results in 6,000 image and text pairs per object category, which is tiny as compared to the 15 million images in the YFCC100M~\cite{thomee2016yfcc100m} data set for training the original CLIP. 

\begin{table}
\setlength{\tabcolsep}{3.8pt} 
\small
\begin{tabular}{lccccccc} 
 \toprule
 Model & \textbf{front} & \textbf{back} & \textbf{left} & \textbf{right} & \textbf{top} & \textbf{bottom} & \\ 
 \midrule
 PRE-TR & 4.12 &  4.09 &  4.12 &  4.12 &  4.09 & 4.15 & \\ 
 FT & 3.91 &  3.90 &  3.91 &  3.89 &  3.97 &   3.92 & \raisebox{-.05\normalbaselineskip}[0pt][0pt]{\rotatebox[origin=c]{90}{\textbf{car}}} \\
 RC-HNS & 2.85 &  2.88 &  3.26 &  2.99 &  3.43 & 3.24 & \\
  \midrule
 PRE-TR & 4.12 &  4.10 &  4.13 &  4.15 &  4.08 & 4.08 & \\ 
 FT & 3.92 &  3.97 &  4.03 &  3.95 &  4.02 &   4.02 & \raisebox{-.05\normalbaselineskip}[0pt][0pt]{\rotatebox[origin=c]{90}{\textbf{airpln}}} \\
 RC-HNS & 3.43 &  3.73 &  3.43 &  3.58 &  3.52 & 3.63 & \\
 \midrule
 PRE-TR & 4.08 &  4.09 &  4.12 &  4.12 &  4.21 & 4.20 & \\ 
 FT & 3.98 &  3.89 &  3.94 &  3.94 &  4.04 &   3.85 & \raisebox{-.05\normalbaselineskip}[0pt][0pt]{\rotatebox[origin=c]{90}{\textbf{mbike}}}\\
 RC-HNS & 2.81 & 2.60 &  2.84 &  2.81 &  3.46 & 3.47 & \\
 \midrule 
 PRE-TR & 4.15 &  4.14 &  4.07 &  4.05 &  4.21 & 4.21 & \\ 
 FT & 3.96 &  3.98 &  3.98 &  3.94 &  3.91 &   3.90 & \raisebox{-.05\normalbaselineskip}[0pt][0pt]{\rotatebox[origin=c]{90}{\textbf{mug}}}\\
 RC-HNS & 3.34 &  3.10 &  3.19 &  2.52 &  2.52 & 2.11 & \\
 \midrule
 PRE-TR & 4.08 &  4.09 &  4.17 &  4.17 &  4.15 & 4.13 & \\ 
 FT & 3.94 &  3.90 &  4.00 &  4.04 &  3.98 &   3.93 & \raisebox{-.05\normalbaselineskip}[0pt][0pt]{\rotatebox[origin=c]{90}{\textbf{bench}}} \\
 RC-HNS & 1.88 &  1.98 &  2.62 &  2.18 &  3.25 &  3.19 & \\
 \bottomrule
\end{tabular}
\caption{KL-Divergence between gold and predicted  viewpoint distribution for the models \textit{PRE-TR, FT, RC-HNS} on the objects \textit{car, airplane, motorbike, mug, bench} for \textit{front, back, left, right, top, bottom} viewpoints on synthetic images. Lower values are better.} 
\label{tab:scoring_function_kullback_leibler_divergence}
\end{table}

\paragraph{Test Set.} For evaluating the retrieval quality for each object category we randomly select three 3D shapes from the ShapeNet test set. 
Then we compute the normalized score distribution on synthetic images around the sphere with radius five for all selected objects of a category, compute the KL-Divergence and average the results per viewpoint query (see Table~\ref{tab:scoring_function_kullback_leibler_divergence}). To assess the performance on real-world data, we carefully curated a data set of 600 images (5 categories $\times$ 6 viewpoints $\times$ 20 images) by retrieving visually similar images for a seed image using image similarity on LAION-5B. Synthetic gold standard views are obtained from the sampled spheres (see Table~\ref{tab:precision_and_recall_metrics}).



\paragraph{Models.} From the official CLIP repository~\cite{openai}, we select ResNet-101~\cite{he2016deep} pre-trained on ImageNet~\cite{deng2009imagenet} as image encoder and pre-trained  BERT model~\cite{devlin2018bert} as query encoder. We compare the following models: (i) \textbf{PRE-TR}ained CLIP, without further fine-tuning, (ii) CLIP-\textbf{FT}, a version of CLIP fine-tuned on the training data with standard cross-entropy loss, (iii) CLIP-\textbf{RC-HNS}, fine-tuned with extended loss objectives explained in Section \ref{sec:model_architecture}.

\subsection{Viewpoint Quality Results}
\label{sec:quality}
Table~\ref{tab:scoring_function_kullback_leibler_divergence} shows the results for the quality of viewpoint retrieval with different models, objects, and viewpoints. 
We find that a pre-trained CLIP model shows a high divergence from the gold standard distribution for all object categories under investigation.  
The fine-tuned model performs slightly better, but still shows large differences from the gold standard. 
The use of random contrasting and hard negative sampling brings the score distribution closer to the gold standard distribution. 
This shows that standard CLIP pre-training and fine-tuning on human-centered 2D images do not produce a suitable scoring function for the viewpoint space around a 3D object.

Evaluating performance on real data using KL divergence is not possible in a similar way as on synthetic data because we do not have access to images from arbitrary viewpoints. Therefore, we compare precision@k and recall@k between synthetic images from ShapeNet and real images at the gold standard viewpoints in Table~\ref{tab:precision_and_recall_metrics}. 
The results show that pre-trained CLIP performs poorly in grounding viewpoints on both synthetic data and real data. Fine-tuning the model on synthetic data greatly improves the retrieval metrics for both synthetic and real data. RC-HNS performs well on synthetic data that is within the distribution, however, it yields slightly lower scores on real-world data in comparison to FT. 
This may result from the fact that
RC-HNS forces the model to generally score out-of-distribution data lower, thereby making the scoring function more sensitive to differences between synthetic and real-world images.
In traditional 2D benchmarks, this may seem like a disadvantage compared to FT, but it proves to be advantageous in 3D viewpoint search, as demonstrated in the following section. Here, the FT model loses performance due to unpredictable scoring behavior in regions far from the gold standard viewpoints. 

\begin{table}
\small  
\setlength{\tabcolsep}{3.2pt} 
\begin{tabular}{l| ccc | cccc} 
 \toprule
 Model &  \textbf{P@1} & \textbf{P@5} & \textbf{P@10} & \textbf{R@1} & \textbf{R@5} & \textbf{R@10}\\ 
 \midrule
 PRE-TR & 0.044 & 0.044 & 0.031 & 0.007 & 0.032 & 0.043 &\\
FT & 0.622 & 0.442 & 0.401 & 0.090 & 0.267 & 0.412 & \raisebox{-.05\normalbaselineskip}[0pt][0pt]{\rotatebox[origin=c]{90}{\textbf{synth}}} \\
RC-HNS & 0.811 & 0.607 & 0.541 & 0.117 & 0.355 & 0.524 & \\ 
\midrule
 PRE-TR & 0.300 & 0.307 & 0.290 & 0.015 & 0.077 & 0.145 & \\ 
 FT & 0.867 & 0.787 & 0.710 & 0.043 & 0.197 & 0.356 & \raisebox{-.05\normalbaselineskip}[0pt][0pt]{\rotatebox[origin=c]{90}{\textbf{real}}} \\
 RC-HNS & 0.733 & 0.673 & 0.633 & 0.036 & 0.168 & 0.317 & \\
 \midrule
\end{tabular}
\caption{Precision@K and Recall@K per model ablation split by synthetic data and real data measured across all object categories.}
\label{tab:precision_and_recall_metrics}
\end{table}

\begin{table}
\small  
\setlength{\tabcolsep}{3.1pt} 
\begin{tabular}{lccccccc} 
 \toprule
 Model &  \textbf{front} & \textbf{back} & \textbf{left} & \textbf{right} & \textbf{top} & \textbf{bottom}\\ 
 \midrule
 PRE-TR & 171.6 & 168.3 & 165.7 & 159.8 & 174.1 & 165.0 &\\
FT & 135.1 & 137.1 & 189.1 & 130.1 & 142.2 & 127.4 & \raisebox{-.05\normalbaselineskip}[0pt][0pt]{\rotatebox[origin=c]{90}{\textbf{Greedy}}} \\
 RC-HNS & 130.5 & 134.5 & 182.7 & 115.9 & 140.3 & 144.4 & \\ 
\midrule
 PRE-TR & 259.4 & 223.2 & 294.0 & 264.8 & 198.4 & 261.6 & \\ 
 FT & 82.4 & 79.1 & 133.0 & 101.1 & 29.7 & 21.5 & \raisebox{-.05\normalbaselineskip}[0pt][0pt]{\rotatebox[origin=c]{90}{\textbf{Bayes}}} \\
 RC-HNS & 73.5 & 62.7 & 62.6 & 49.4 & 22.0 & 22.9 & \\
 \midrule
\end{tabular}
\caption{Average number of calls to the scoring function per search algorithm and viewpoint query.}
\label{tab:search_algorith_performance_comparison}
\end{table}

\begin{figure*}[ht]
 \centering 
 \includegraphics[width=\linewidth]{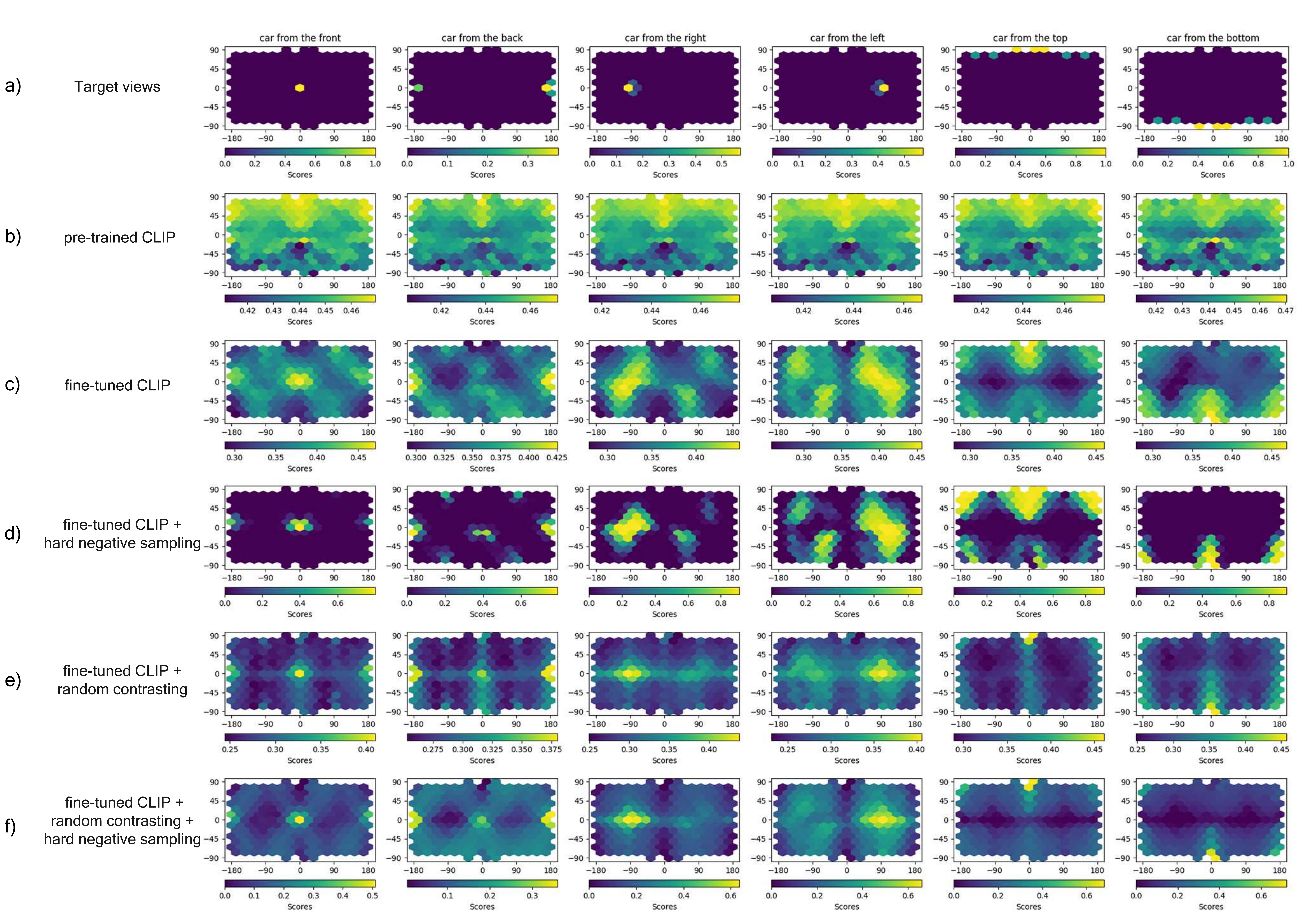}
 \caption{Score distribution on the six viewpoints per loss function combination on a car object. In a)  gold-standard viewpoints expected to have high scores are shown, b) pre-trained CLIP, c) fine-tuned CLIP, d) hard negative sampling, e) random contrasting, and f) random contrasting + hard negative sampling. For more, see appendix~\ref{sec:appendix_a}.}
 \label{fig:score_functions_overview}
\end{figure*}

\subsection{Search Performance Results}
\label{sec:search_performance_results}
We test search performance in 3D as described in Section~\ref{sec:metrics} for all six queries. Table~\ref{tab:search_algorith_performance_comparison} illustrates the performance for Greedy and Bayes search.
Both algorithms perform significantly better than an exhaustive search on the Goldberg polyhedron (= 1002 sample points, fixed radius). 
Bayesian search is much faster than greedy search, when using a finetuned scoring function (FT, RC-HNS), and it is more affected by the shape of the scoring function since it samples it strategically: it is fastest with the smoothest scoring function RC-HNS and very slow with pretrained CLIP. This is in line with the viewpoint quality results in Section \ref{sec:quality}, showing that pretrained CLIP has a poor representation of the viewpoint space around an object.

\section{Analysis}
\label{sec:analysis}
This section takes a closer look at how well the text-viewpoint embeddings capture understanding of different viewpoints. Specifically, we will explore whether the scoring functions correctly identify viewpoints that align with the linguistic description, while providing lower scores for those that do not.

\subsection{Exhaustive Viewpoint Space Analysis}
Based on the polyhedron that defines the viewpoint space of the camera, we carry out an exhaustive analysis of the scoring function over this space for specific objects and queries. We select a car from the test set of the ShapeNet data set and plot the scores of the evenly distributed samples from the surface of the Goldberg polyhedron at a radius of five for the six canonical viewpoint queries.
We examine five different configurations of the loss objective shown in Equation (4). Figure~\ref{fig:score_functions_overview}a) illustrates the target region on the hexagon diagram, which contains the optimal viewpoint for a given query.  
It can be seen in Figure~\ref{fig:score_functions_overview}b) that a pre-trained CLIP model even if trained on a large data set, is not able to discriminate between different viewpoints and that the scoring function has multiple optima. 
Fine-tuning the CLIP model (\ref{fig:score_functions_overview}c) on synthetic images improves viewpoint discriminability. Nevertheless, apart from the absolute gold standard regions, the function shows problematic local optima and in particular the left and right side views of the car are difficult to distinguish. 
In (d), we fine-tune the CLIP model by applying the hard negative sampling strategy proposed by~\citet{robinson2020contrastive}. The results show that the gold standard viewpoints can be distinguished much more effectively when compared to previous experiments. However, the transition between viewpoints is quite sudden, making it challenging for a search algorithm to reach the optimum. 
In (e), 
a combination of negative contrastive loss $L_{\mathbf{v},\mathbf{q}}$ and random contrastive loss $L_{r}$  is applied. 
The results show that the additional objective makes the scoring function much more stable in regions farther away from known canonical viewpoints.
In experiment (f), we combine hard negative sampling $L_{h}$ with the idea of random contrasting. 
The plot of the scoring function shows that for each canonical viewpoint, the function increases steadily toward the optimal view.

\subsection{Nonsensical Viewpoints}
\label{sec:nonsensical_viewpoints}
\begin{figure}[h]
 \centering 
 \includegraphics[width=\columnwidth]{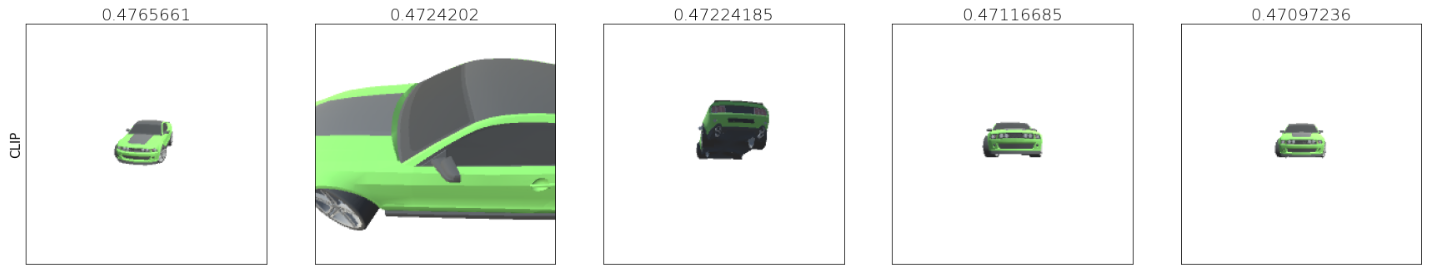}
 \caption{Retrieved nonsensical viewpoints in the top-5 scored images on CLIP for the query \textit{a picture of a car from the front}.}
 \label{fig:nonsensical_viewpoints_figure}
\end{figure}

A further problem we noticed is that CLIP predicts high scores for nonsensical views that do not relate to  the query, but rather seem to activate certain features to drive up the score, similar to adversarial examples~\cite{goodfellow2014explaining}. Such behavior of models on unseen images has also been described by~\citet{du2022vos} and should be considered when using CLIP representations in continuous 3D environments, especially for vision-and-language navigation tasks, as in~\citet{khandelwal2022simple}. Figure~\ref{fig:nonsensical_viewpoints_figure} shows retrieved nonsensical viewpoint images among the top-5 for \textit{car from the front}. 

\subsection{Data Set Size Ablations}
\label{sec:section_low_training_data_ablations}
\begin{figure}[h]
 \centering 
 \includegraphics[width=\columnwidth]{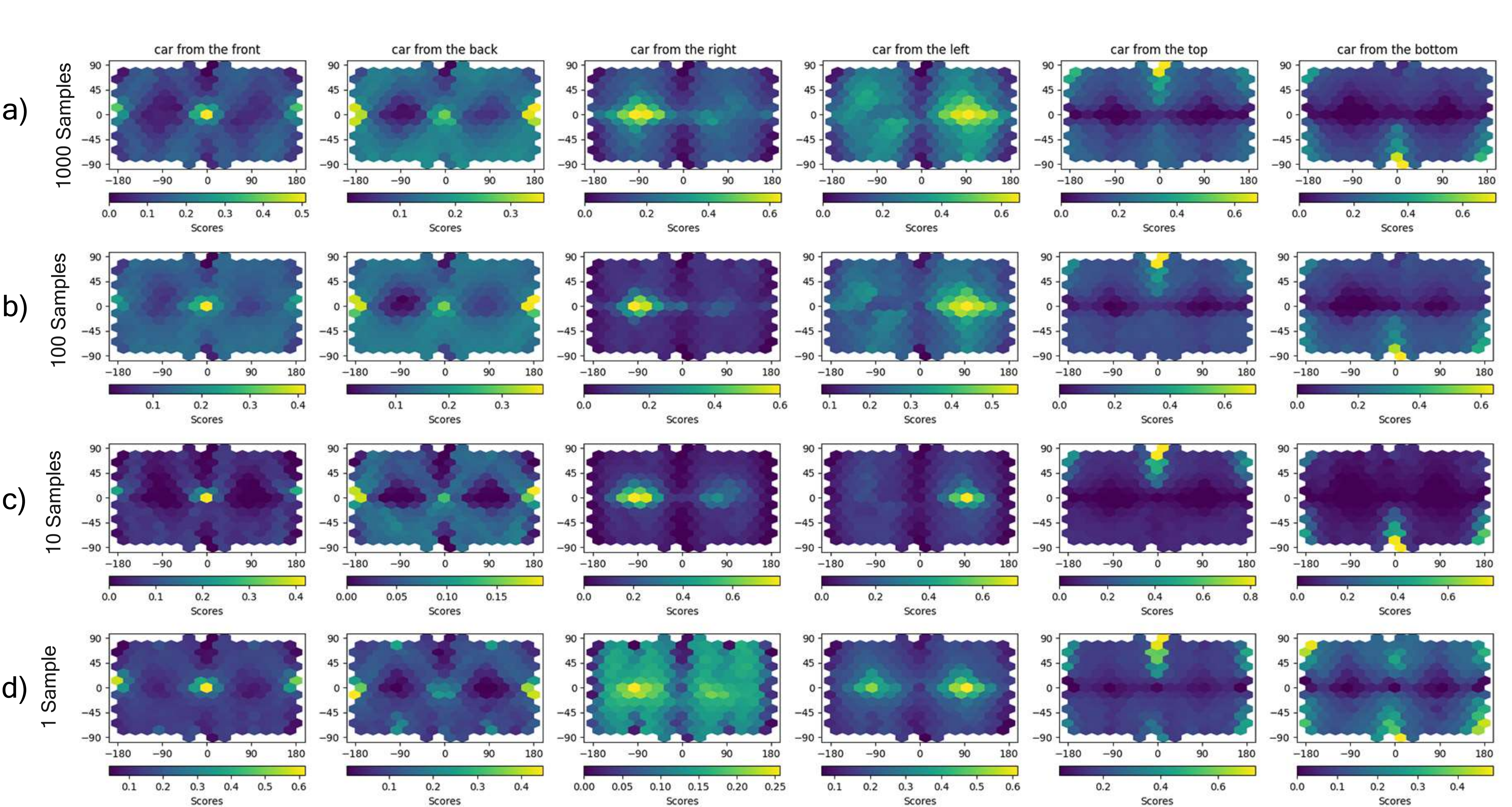}
 \caption{Overview of the effects of gradually reducing the number of training images per view from a) 1000 to b) 100 to c) 10 to d) 1 on CLIP-RC-HNS.}
\label{fig:low_data_ablation_experiment_figure}
\end{figure}

To test how the scoring function is affected when only a small amount of training data is available, we gradually reduce the number of training samples from 1,000 to 1 for the best-performing model CLIP-RC-HNS.
Access to 1,000 training examples per viewpoint, as shown in \ref{fig:low_data_ablation_experiment_figure}a), leads to a smooth function. 
Reducing the training data by 90 percent to 100 examples per viewpoint keeps  good performance for the target viewpoints. Compared to the full data set, smoothness suffers slightly. 
Reducing the training data by 99 percent to ten samples per viewpoint still allows good results in the target regions. However, the surrounding regions become less smooth and drop more abruptly. 
Surprisingly, when breaking down the training data to one example per viewpoint, the target viewpoint areas still lead to global optima in all search queries. However, the transitions are no longer smooth but rather abrupt, especially for the front and back. 

\section{Conclusion}
\label{sec:conclusion}
We developed a new framework to assess the capabilities of L\&V models to ground viewpoint descriptions. 
Through our research, we discovered that a standard CLIP model struggles to distinguish between different viewpoints. 
To address this, we explored a combination of different loss objectives on synthetic data to make it easier to retrieve viewpoints from language descriptions. 
Our experiments revealed that incorporating random contrasting leads to a more accurate and seamless scoring function, as compared to using only text and human-centric images. 
Our framework thus offers a promising approach to scale L\&V models trained on large-scale image-text datasets for applications that involve interaction in the 3D world.

\newpage

\subsection*{Limitations}
\label{sec:limitations}
We deliberately opted for a simple controllable setup in order to gain a precise understanding of viewpoint representation in CLIP. 
Our experiments are restricted to canonical views and canned descriptions since they are easy to generate and evaluate automatically. 
Extending the data to other views and to human-like descriptions is the obvious avenue for future research. 
In particular, with the advent of NERF models in computer vision, we look forward to integrating these types of models into our framework, as this would allow the generation of near-realistic images in a controlled 3D setup, which would allow for even better evaluation of scoring functions in text viewpoint retrieval. 
Varying the level of detail of the 3D shapes, especially in complex 3D scenes where large objects consist of smaller parts is another interesting direction. 
Another restriction of our set-up is the fact that we consider context-free retrieval of viewpoints, whereas in many human-like descriptions such as the \textit{right front tire of a car}, the viewpoint may not be visually unique and depend on the context of the scene, such as the relative position of the viewpoint to other viewpoints.
The same applies to views that need to be delivered to a user in a task-oriented interaction, and are likely to be more complex and diverse than the canonical and synthetic ones used in this work. 
In conclusion, we believe that our framework has the potential to provide a more comprehensive understanding of reporting biases in image-text data used for pre-training L\&V models. By conducting a 360-degree analysis of the scoring function, our framework allows for a more thorough examination of these biases, as everything is visible and nothing can be hidden from the investigator, unlike when evaluating against a set of gold-standard viewpoints.

\section*{Ethics Statement}
3D models from the ShapeNet dataset are available for research and non-commercial purposes as well as the LAION-5B data set. 
We did not collect any personal information from any annotators.
We clearly state the intended use of our models, which is to support human-centric interaction with AI models in the 3D world.


\section*{Acknowledgments}
We thank the Michael Stifel Center Jena for funding this work, which is part of the Carl Zeiss Foundation-funded project 'A Virtual Workshop for Digitization in the Sciences' (062017-02).

\bibliography{custom}

\begin{thebibliography}{44}
\expandafter\ifx\csname natexlab\endcsname\relax\def\natexlab#1{#1}\fi

\bibitem[{Achlioptas et~al.(2019)Achlioptas, Fan, Hawkins, Goodman, and
  Guibas}]{achlioptas2019shapeglot}
Panos Achlioptas, Judy Fan, Robert Hawkins, Noah Goodman, and Leonidas~J
  Guibas. 2019.
\newblock Shapeglot: Learning language for shape differentiation.
\newblock In \emph{Proceedings of the IEEE/CVF International Conference on
  Computer Vision}, pages 8938--8947.

\bibitem[{Arbel and Ferrie(1999)}]{arbel1999viewpoint}
Tal Arbel and Frank~P Ferrie. 1999.
\newblock Viewpoint selection by navigation through entropy maps.
\newblock In \emph{Proceedings of the Seventh IEEE International Conference on
  Computer Vision}, volume~1, pages 248--254. IEEE.

\bibitem[{Baldrati et~al.(2022)Baldrati, Bertini, Uricchio, and
  Del~Bimbo}]{baldrati2022effective}
Alberto Baldrati, Marco Bertini, Tiberio Uricchio, and Alberto Del~Bimbo. 2022.
\newblock Effective conditioned and composed image retrieval combining
  clip-based features.
\newblock In \emph{Proceedings of the IEEE/CVF Conference on Computer Vision
  and Pattern Recognition}, pages 21466--21474.

\bibitem[{Bonaventura et~al.(2018)Bonaventura, Feixas, Sbert, Chuang, and
  Wallraven}]{bonaventura2018survey}
Xavier Bonaventura, Miquel Feixas, Mateu Sbert, Lewis Chuang, and Christian
  Wallraven. 2018.
\newblock A survey of viewpoint selection methods for polygonal models.
\newblock \emph{Entropy}, 20(5):370.

\bibitem[{Bordoloi and Shen(2005)}]{bordoloi2005view}
Udeepta~D Bordoloi and H-W Shen. 2005.
\newblock View selection for volume rendering.
\newblock In \emph{VIS 05. IEEE Visualization, 2005.}, pages 487--494. IEEE.

\bibitem[{Chang et~al.(2015)Chang, Funkhouser, Guibas, Hanrahan, Huang, Li,
  Savarese, Savva, Song, Su et~al.}]{chang2015shapenet}
Angel~X Chang, Thomas Funkhouser, Leonidas Guibas, Pat Hanrahan, Qixing Huang,
  Zimo Li, Silvio Savarese, Manolis Savva, Shuran Song, Hao Su, et~al. 2015.
\newblock Shapenet: An information-rich 3d model repository.
\newblock \emph{arXiv preprint arXiv:1512.03012}.

\bibitem[{Deng et~al.(2009)Deng, Dong, Socher, Li, Li, and
  Fei-Fei}]{deng2009imagenet}
Jia Deng, Wei Dong, Richard Socher, Li-Jia Li, Kai Li, and Li~Fei-Fei. 2009.
\newblock Imagenet: A large-scale hierarchical image database.
\newblock In \emph{2009 IEEE conference on computer vision and pattern
  recognition}, pages 248--255. Ieee.

\bibitem[{Devlin et~al.(2018)Devlin, Chang, Lee, and
  Toutanova}]{devlin2018bert}
Jacob Devlin, Ming-Wei Chang, Kenton Lee, and Kristina Toutanova. 2018.
\newblock Bert: Pre-training of deep bidirectional transformers for language
  understanding.
\newblock \emph{arXiv preprint arXiv:1810.04805}.

\bibitem[{Du et~al.(2022)Du, Wang, Cai, and Li}]{du2022vos}
Xuefeng Du, Zhaoning Wang, Mu~Cai, and Yixuan Li. 2022.
\newblock Vos: Learning what you don't know by virtual outlier synthesis.
\newblock \emph{arXiv preprint arXiv:2202.01197}.

\bibitem[{Fan et~al.(2022)Fan, Chen, Jiang, Zhou, Zhang, and
  Wang}]{fan2022aerial}
Yue Fan, Winson Chen, Tongzhou Jiang, Chun Zhou, Yi~Zhang, and Xin~Eric Wang.
  2022.
\newblock Aerial vision-and-dialog navigation.
\newblock \emph{arXiv preprint arXiv:2205.12219}.

\bibitem[{Fang et~al.(2021)Fang, Xiong, Xu, and Chen}]{fang2021clip2video}
Han Fang, Pengfei Xiong, Luhui Xu, and Yu~Chen. 2021.
\newblock Clip2video: Mastering video-text retrieval via image clip.
\newblock \emph{arXiv preprint arXiv:2106.11097}.

\bibitem[{Gadre et~al.(2022)Gadre, Wortsman, Ilharco, Schmidt, and
  Song}]{gadre2022clip}
Samir~Yitzhak Gadre, Mitchell Wortsman, Gabriel Ilharco, Ludwig Schmidt, and
  Shuran Song. 2022.
\newblock Clip on wheels: Zero-shot object navigation as object localization
  and exploration.
\newblock \emph{arXiv preprint arXiv:2203.10421}.

\bibitem[{Goldberg(1937)}]{goldberg1937class}
Michael Goldberg. 1937.
\newblock A class of multi-symmetric polyhedra.
\newblock \emph{Tohoku Mathematical Journal, First Series}, 43:104--108.

\bibitem[{Goodfellow et~al.(2014)Goodfellow, Shlens, and
  Szegedy}]{goodfellow2014explaining}
Ian~J Goodfellow, Jonathon Shlens, and Christian Szegedy. 2014.
\newblock Explaining and harnessing adversarial examples.
\newblock \emph{arXiv preprint arXiv:1412.6572}.

\bibitem[{He et~al.(2016)He, Zhang, Ren, and Sun}]{he2016deep}
Kaiming He, Xiangyu Zhang, Shaoqing Ren, and Jian Sun. 2016.
\newblock Deep residual learning for image recognition.
\newblock In \emph{Proceedings of the IEEE conference on computer vision and
  pattern recognition}, pages 770--778.

\bibitem[{Head et~al.(2021)Head, Kumar, Nahrstaedt, Louppe, and
  Shcherbatyi}]{https://doi.org/10.5281/zenodo.1157319}
Tim Head, Manoj Kumar, Holger Nahrstaedt, Gilles Louppe, and Iaroslav
  Shcherbatyi. 2021.
\newblock \href {https://doi.org/10.5281/ZENODO.1157319}
  {scikit-optimize/scikit-optimize}.

\bibitem[{Kamada and Kawai(1988)}]{kamada1988simple}
Tomihisa Kamada and Satoru Kawai. 1988.
\newblock A simple method for computing general position in displaying
  three-dimensional objects.
\newblock \emph{Computer Vision, Graphics, and Image Processing}, 41(1):43--56.

\bibitem[{Khandelwal et~al.(2022)Khandelwal, Weihs, Mottaghi, and
  Kembhavi}]{khandelwal2022simple}
Apoorv Khandelwal, Luca Weihs, Roozbeh Mottaghi, and Aniruddha Kembhavi. 2022.
\newblock Simple but effective: Clip embeddings for embodied ai.
\newblock In \emph{Proceedings of the IEEE/CVF Conference on Computer Vision
  and Pattern Recognition}, pages 14829--14838.

\bibitem[{Kullback and Leibler(1951)}]{kullback1951information}
Solomon Kullback and Richard~A Leibler. 1951.
\newblock On information and sufficiency.
\newblock \emph{The annals of mathematical statistics}, 22(1):79--86.

\bibitem[{Lee et~al.(2011)Lee, Mishchenko, Shen, and Crawfis}]{lee2011view}
Teng-Yok Lee, Oleg Mishchenko, Han-Wei Shen, and Roger Crawfis. 2011.
\newblock View point evaluation and streamline filtering for flow
  visualization.
\newblock In \emph{2011 IEEE Pacific Visualization Symposium}, pages 83--90.
  IEEE.

\bibitem[{Leifman et~al.(2016)Leifman, Shtrom, and Tal}]{leifman2016surface}
George Leifman, Elizabeth Shtrom, and Ayellet Tal. 2016.
\newblock Surface regions of interest for viewpoint selection.
\newblock \emph{IEEE transactions on pattern analysis and machine
  intelligence}, 38(12):2544--2556.

\bibitem[{Li et~al.(2020{\natexlab{a}})Li, Duan, Fang, Gong, and
  Jiang}]{li2020unicoder}
Gen Li, Nan Duan, Yuejian Fang, Ming Gong, and Daxin Jiang. 2020{\natexlab{a}}.
\newblock Unicoder-vl: A universal encoder for vision and language by
  cross-modal pre-training.
\newblock In \emph{Proceedings of the AAAI Conference on Artificial
  Intelligence}, volume~34, pages 11336--11344.

\bibitem[{Li et~al.(2021)Li, Selvaraju, Gotmare, Joty, Xiong, and
  Hoi}]{li2021align}
Junnan Li, Ramprasaath Selvaraju, Akhilesh Gotmare, Shafiq Joty, Caiming Xiong,
  and Steven Chu~Hong Hoi. 2021.
\newblock Align before fuse: Vision and language representation learning with
  momentum distillation.
\newblock \emph{Advances in Neural Information Processing Systems}, 34.

\bibitem[{Li et~al.(2020{\natexlab{b}})Li, Yin, Li, Zhang, Hu, Zhang, Wang, Hu,
  Dong, Wei et~al.}]{li2020oscar}
Xiujun Li, Xi~Yin, Chunyuan Li, Pengchuan Zhang, Xiaowei Hu, Lei Zhang, Lijuan
  Wang, Houdong Hu, Li~Dong, Furu Wei, et~al. 2020{\natexlab{b}}.
\newblock Oscar: Object-semantics aligned pre-training for vision-language
  tasks.
\newblock In \emph{European Conference on Computer Vision}, pages 121--137.
  Springer.

\bibitem[{Meuschke et~al.(2017)Meuschke, Engelke, Beuing, Preim, and
  Lawonn}]{meuschke2017automatic}
Monique Meuschke, Wito Engelke, Oliver Beuing, Bernhard Preim, and Kai Lawonn.
  2017.
\newblock Automatic viewpoint selection for exploration of time-dependent
  cerebral aneurysm data.
\newblock In \emph{Bildverarbeitung fuer die Medizin 2017}, pages 352--357.
  Springer.

\bibitem[{Mockus(1994)}]{mockus1994application}
Jonas Mockus. 1994.
\newblock Application of bayesian approach to numerical methods of global and
  stochastic optimization.
\newblock \emph{Journal of Global Optimization}, 4(4):347--365.

\bibitem[{Mokady et~al.(2021)Mokady, Hertz, and Bermano}]{mokady2021clipcap}
Ron Mokady, Amir Hertz, and Amit~H Bermano. 2021.
\newblock Clipcap: Clip prefix for image captioning.
\newblock \emph{arXiv preprint arXiv:2111.09734}.

\bibitem[{M{\"u}hler et~al.(2007)M{\"u}hler, Neugebauer, Tietjen, and
  Preim}]{muhler2007viewpoint}
Konrad M{\"u}hler, Mathias Neugebauer, Christian Tietjen, and Bernhard Preim.
  2007.
\newblock Viewpoint selection for intervention planning.
\newblock In \emph{EuroVis}, pages 267--274.

\bibitem[{Neugebauer et~al.(2013)Neugebauer, Lawonn, Beuing, Berg, Janiga, and
  Preim}]{neugebauer2013amnivis}
Mathias Neugebauer, Kai Lawonn, Oliver Beuing, Philipp Berg, Gabor Janiga, and
  Bernhard Preim. 2013.
\newblock Amnivis--a system for qualitative exploration of near-wall
  hemodynamics in cerebral aneurysms.
\newblock In \emph{Computer Graphics Forum}, volume~32, pages 251--260. Wiley
  Online Library.

\bibitem[{Olague and Mohr(2002)}]{olague2002optimal}
Gustavo Olague and Roger Mohr. 2002.
\newblock Optimal camera placement for accurate reconstruction.
\newblock \emph{Pattern recognition}, 35(4):927--944.

\bibitem[{OpenAI()}]{openai}
OpenAI.
\newblock \href {https://github.com/openai/CLIP} {Openai/clip: Contrastive
  language-image pretraining}.

\bibitem[{Plemenos and Sokolov(2006)}]{plemenos2006viewpoint}
Dimitri Plemenos and Dmitry Sokolov. 2006.
\newblock Viewpoint quality and scene understanding.
\newblock In \emph{Eurographics Symposium on Virtual Reality}, pages 67--73.
  VAST'2005.

\bibitem[{Podolak et~al.(2006)Podolak, Shilane, Golovinskiy, Rusinkiewicz, and
  Funkhouser}]{podolak2006planar}
Joshua Podolak, Philip Shilane, Aleksey Golovinskiy, Szymon Rusinkiewicz, and
  Thomas Funkhouser. 2006.
\newblock A planar-reflective symmetry transform for 3d shapes.
\newblock In \emph{ACM SIGGRAPH 2006 Papers}, pages 549--559.

\bibitem[{Radford et~al.(2021)Radford, Kim, Hallacy, Ramesh, Goh, Agarwal,
  Sastry, Askell, Mishkin, Clark et~al.}]{radford2021learning}
Alec Radford, Jong~Wook Kim, Chris Hallacy, Aditya Ramesh, Gabriel Goh,
  Sandhini Agarwal, Girish Sastry, Amanda Askell, Pamela Mishkin, Jack Clark,
  et~al. 2021.
\newblock Learning transferable visual models from natural language
  supervision.
\newblock In \emph{International Conference on Machine Learning}, pages
  8748--8763. PMLR.

\bibitem[{Roberts and Marshall(1998)}]{roberts1998viewpoint}
DR~Roberts and A~David Marshall. 1998.
\newblock Viewpoint selection for complete surface coverage of three
  dimensional objects.
\newblock In \emph{BMVC}, pages 1--11. Citeseer.

\bibitem[{Robinson et~al.(2020)Robinson, Chuang, Sra, and
  Jegelka}]{robinson2020contrastive}
Joshua Robinson, Ching-Yao Chuang, Suvrit Sra, and Stefanie Jegelka. 2020.
\newblock Contrastive learning with hard negative samples.
\newblock \emph{arXiv preprint arXiv:2010.04592}.

\bibitem[{Schuhmann et~al.(2021)Schuhmann, Vencu, Beaumont, Kaczmarczyk,
  Mullis, Katta, Coombes, Jitsev, and Komatsuzaki}]{schuhmann2021laion}
Christoph Schuhmann, Richard Vencu, Romain Beaumont, Robert Kaczmarczyk,
  Clayton Mullis, Aarush Katta, Theo Coombes, Jenia Jitsev, and Aran
  Komatsuzaki. 2021.
\newblock Laion-400m: Open dataset of clip-filtered 400 million image-text
  pairs.
\newblock \emph{arXiv preprint arXiv:2111.02114}.

\bibitem[{Shen et~al.(2021)Shen, Li, Tan, Bansal, Rohrbach, Chang, Yao, and
  Keutzer}]{shen2021much}
Sheng Shen, Liunian~Harold Li, Hao Tan, Mohit Bansal, Anna Rohrbach, Kai-Wei
  Chang, Zhewei Yao, and Kurt Keutzer. 2021.
\newblock How much can clip benefit vision-and-language tasks?
\newblock \emph{arXiv preprint arXiv:2107.06383}.

\bibitem[{Tao et~al.(2012)Tao, Ma, Wang, and Shene}]{tao2012unified}
Jun Tao, Jun Ma, Chaoli Wang, and Ching-Kuang Shene. 2012.
\newblock A unified approach to streamline selection and viewpoint selection
  for 3d flow visualization.
\newblock \emph{IEEE Transactions on Visualization and Computer Graphics},
  19(3):393--406.

\bibitem[{Thomason et~al.(2020)Thomason, Murray, Cakmak, and
  Zettlemoyer}]{thomason2020vision}
Jesse Thomason, Michael Murray, Maya Cakmak, and Luke Zettlemoyer. 2020.
\newblock Vision-and-dialog navigation.
\newblock In \emph{Conference on Robot Learning}, pages 394--406. PMLR.

\bibitem[{Thomason et~al.(2022)Thomason, Shridhar, Bisk, Paxton, and
  Zettlemoyer}]{thomason2022language}
Jesse Thomason, Mohit Shridhar, Yonatan Bisk, Chris Paxton, and Luke
  Zettlemoyer. 2022.
\newblock Language grounding with 3d objects.
\newblock In \emph{Conference on Robot Learning}, pages 1691--1701. PMLR.

\bibitem[{Thomee et~al.(2016)Thomee, Shamma, Friedland, Elizalde, Ni, Poland,
  Borth, and Li}]{thomee2016yfcc100m}
Bart Thomee, David~A Shamma, Gerald Friedland, Benjamin Elizalde, Karl Ni,
  Douglas Poland, Damian Borth, and Li-Jia Li. 2016.
\newblock Yfcc100m: The new data in multimedia research.
\newblock \emph{Communications of the ACM}, 59(2):64--73.

\bibitem[{V{\'a}zquez et~al.(2001)V{\'a}zquez, Feixas, Sbert, and
  Heidrich}]{vazquez2001viewpoint}
Pere-Pau V{\'a}zquez, Miquel Feixas, Mateu Sbert, and Wolfgang Heidrich. 2001.
\newblock Viewpoint selection using viewpoint entropy.
\newblock In \emph{VMV}, volume~1, pages 273--280. Citeseer.

\bibitem[{Yu et~al.(2022)Yu, Wang, Vasudevan, Yeung, Seyedhosseini, and
  Wu}]{yu2022coca}
Jiahui Yu, Zirui Wang, Vijay Vasudevan, Legg Yeung, Mojtaba Seyedhosseini, and
  Yonghui Wu. 2022.
\newblock Coca: Contrastive captioners are image-text foundation models.
\newblock \emph{arXiv preprint arXiv:2205.01917}.

\end{thebibliography}
\bibliographystyle{acl_natbib}

\appendix
\newpage
\clearpage
\section{Experiment Details}
\label{sec:appendix_a}
This section provides additional details on our experimental setup. 
Section~\ref{sec:appendix_scoring_function_analysis} contains further visualizations of the experiments discussed in section~\ref{sec:experiments}. Section~\ref{sec:search_algorithm_analysis_appendix} provides details about the implementation of the search algorithms used in our benchmark.

\subsection{Scoring Function Analysis}
\label{sec:appendix_scoring_function_analysis}
The following plots illustrate the score distributions obtained with the different model ablations CLIP-\textbf{PRE-TR}, CLIP-\textbf{FT}, and CLIP-\textbf{RC-HNS}. 

\begin{figure*}[h]
     \centering
     \begin{subfigure}{1.0\textwidth}
         \centering
         \includegraphics[width=\textwidth]{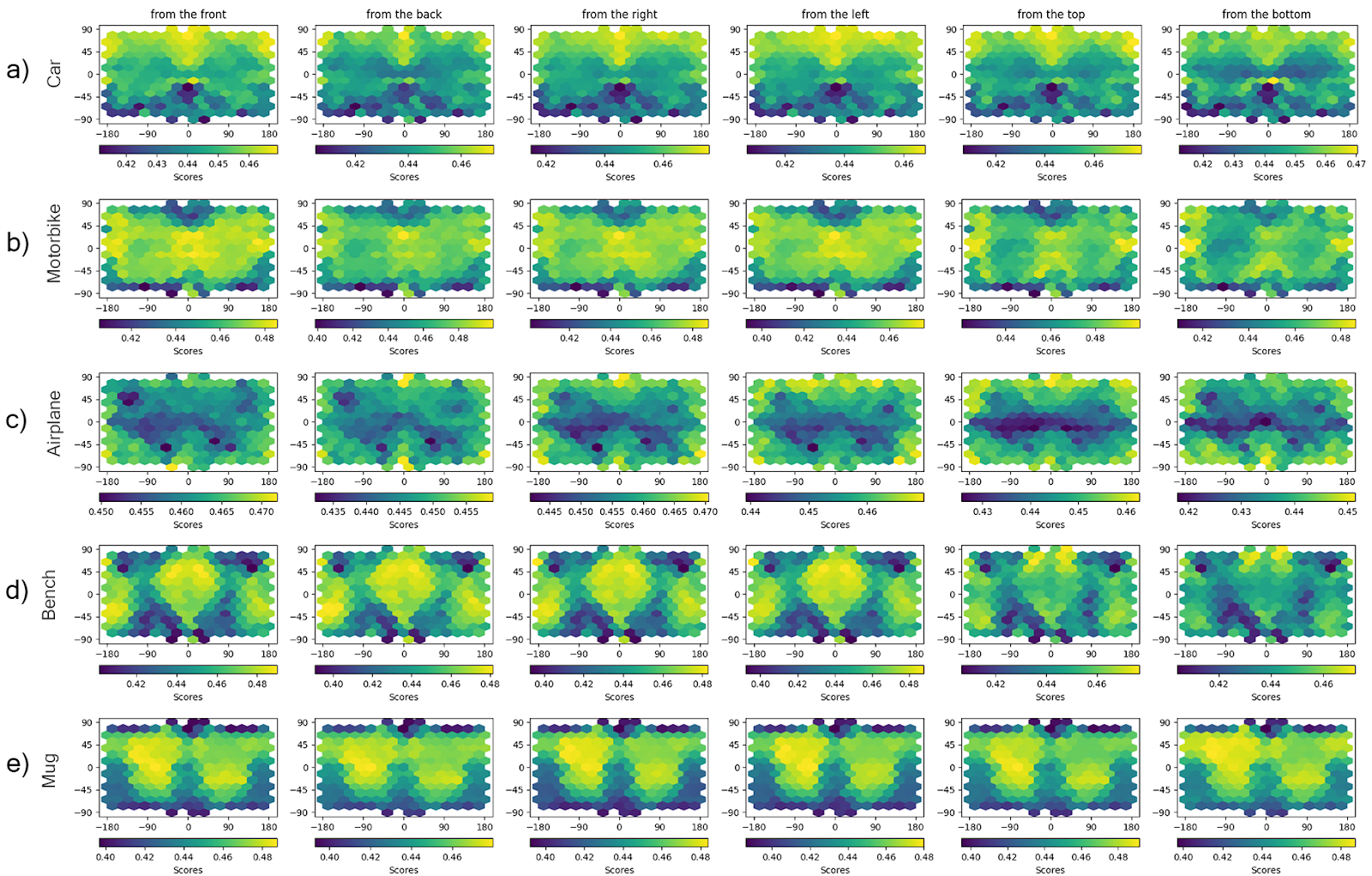}
         \caption{Scoring Function Distribution of CLIP PRE-TR model on cars, motorbikes, airplanes, benches, and mugs for the six canonical viewpoint queries.}
         \label{fig:clip_scoring_function_distribution}
     \end{subfigure}
     \hfill
     \begin{subfigure}{1.0\textwidth}
         \centering
         \includegraphics[width=\textwidth]{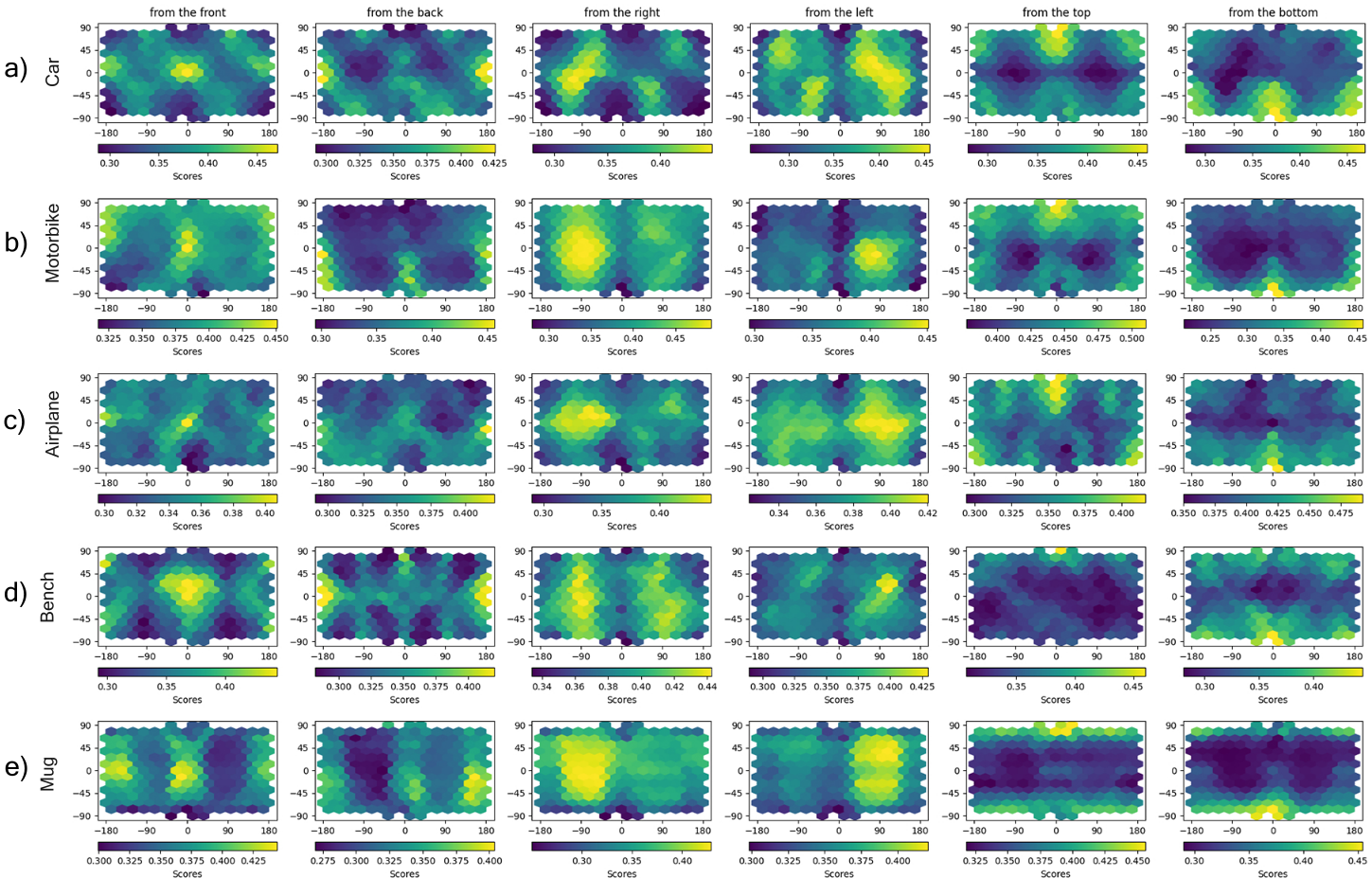}
         \caption{Scoring Function Distribution of the CLIP-FT model on cars, motorbikes, airplanes, benches, and mugs for the six canonical viewpoint queries.}
         \label{fig:clip_ft_scoring_function_distribution}
     \end{subfigure}
     \caption{Scoring Function Distributions on CLIP PRE-TR and CLIP-FT.}
\end{figure*}

\begin{figure*}[h]
     \centering
     \begin{subfigure}{1.0\textwidth}
         \centering
         \includegraphics[width=\textwidth]{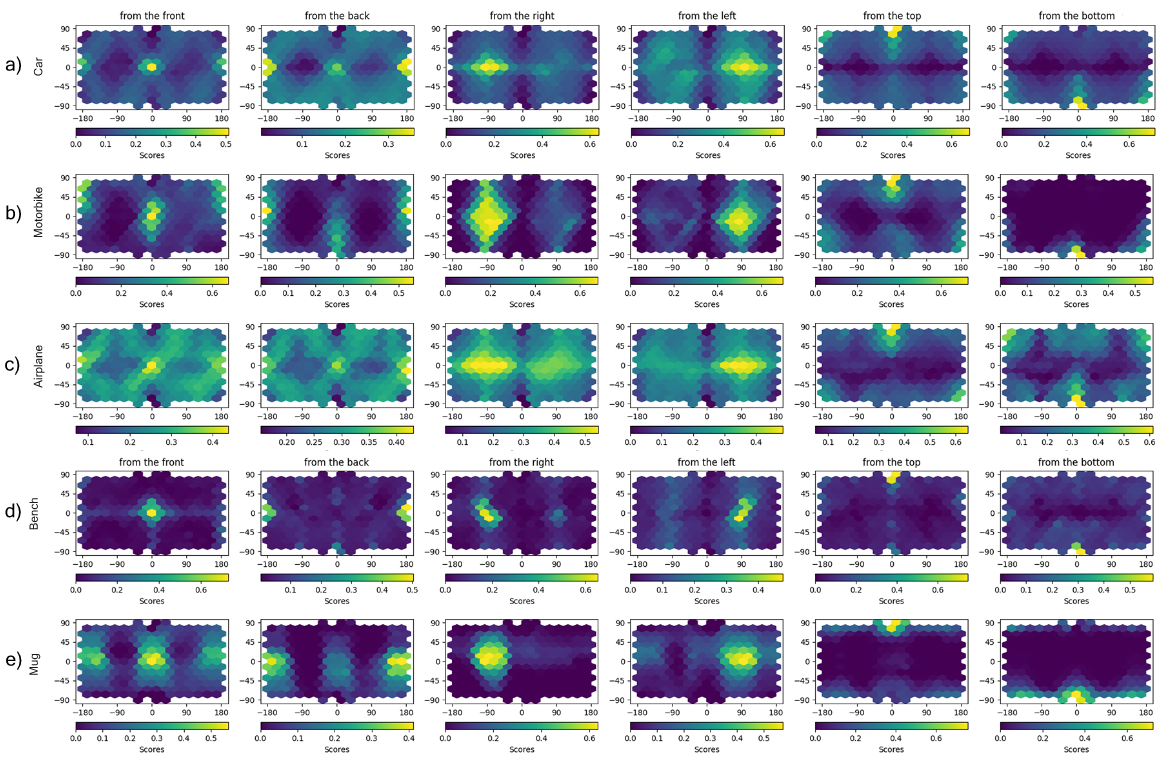}
         \caption{Scoring Function Distribution of the CLIP-RC-HNS model on cars, motorbikes, airplanes, benches, and mugs for the six canonical viewpoint queries.}
         \label{fig:clip_rc_hnm_scoring_function_distribution}
     \end{subfigure}
      \hfill
     \begin{subfigure}{0.8\textwidth}
         \centering
         \includegraphics[width=\textwidth]{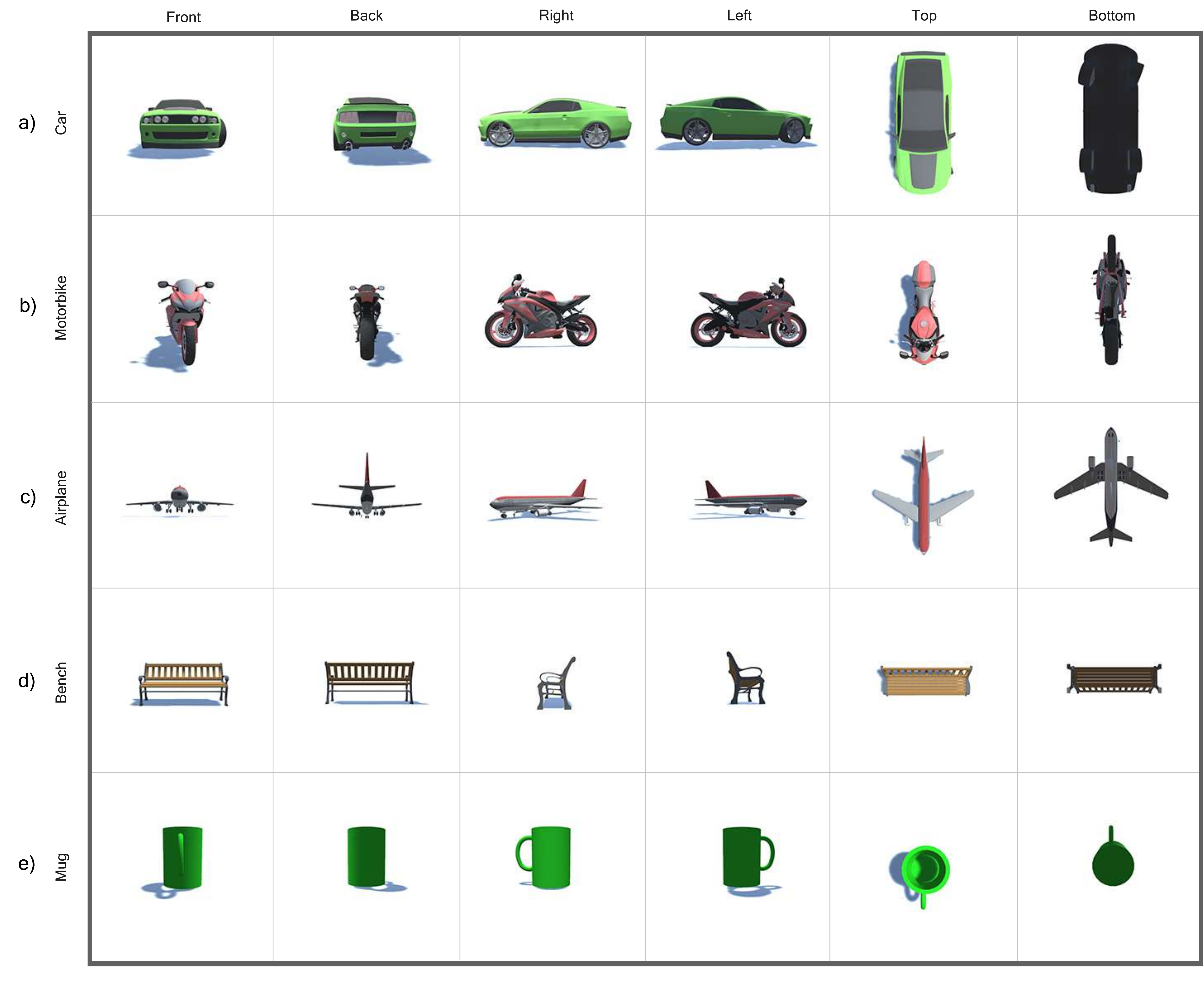}
         \caption{Optimal viewpoints of the six canonical views for a) cars, b) motorbikes, c) airplanes, d) benches, and e) mugs of the ShapeNet data set~\cite{chang2015shapenet} retrieved from the optima of the CLIP-RC-HNS scoring function.}
         \label{fig:optimal_viewpoint_images_overview}
     \end{subfigure}
     \caption{Scoring Function Distributions on CLIP-RC-HNS and retrieved viewpoint images.}
\end{figure*}


\paragraph{Scoring Function PRETR.}
Figure~\ref{fig:clip_scoring_function_distribution} shows the score distribution of the PRE-TRained CLIP model over 3D objects from the test set of the ShapeNet dataset. 

\paragraph{Scoring Function FT.}
Figure~\ref{fig:clip_ft_scoring_function_distribution} depicts the scoring distribution of the CLIP-FT model over 3D objects from the test set of the ShapeNet dataset. 



\paragraph{Scoring Function RC-HNS.}
Figure~\ref{fig:clip_rc_hnm_scoring_function_distribution} illustrates the score distribution of the CLIP-RC-HNS model over 3D objects from the test set of the ShapeNet dataset. 

\begin{figure*}[h]
     \centering
     \begin{subfigure}{0.9\textwidth}
         \centering
         \includegraphics[width=\textwidth]{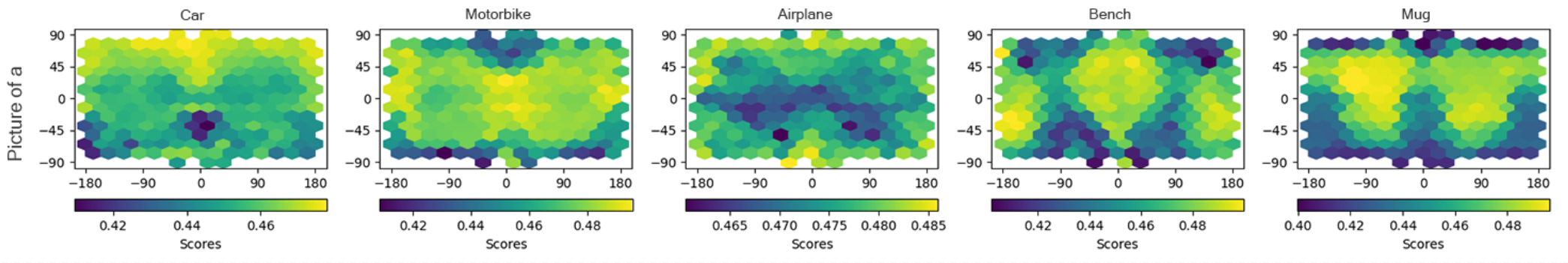}
         \caption{Scoring function distribution on cars, motorbikes, airplanes, benches, and mugs given the query \textit{a picture of an X}, where X stands as a variable for \textit{car/motorbike/airplane/bench/mug}}
         \label{fig:pure_object_queries_comparison}
     \end{subfigure}
     
     \vspace{5cm}
     
     \begin{subfigure}{0.9\textwidth}
         \centering
         \includegraphics[width=\textwidth]{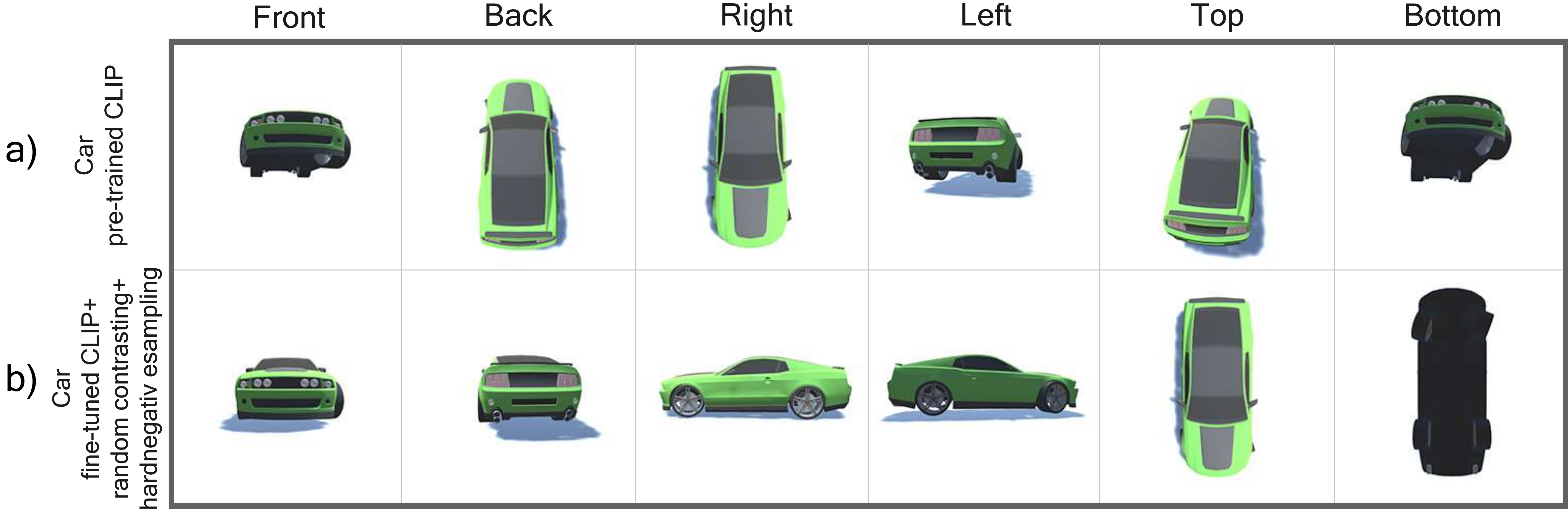}
         \caption{Comparison of optimal viewpoints of the six canonical views between a) PRE-TRained CLIP and b) CLIP-RC-HNS.}
      \label{fig:clip_compare_screenshots_car}
     \end{subfigure}
     
     \vspace{5cm}
     
     \begin{subfigure}{0.9\textwidth}
         \centering
         \includegraphics[width=\textwidth]{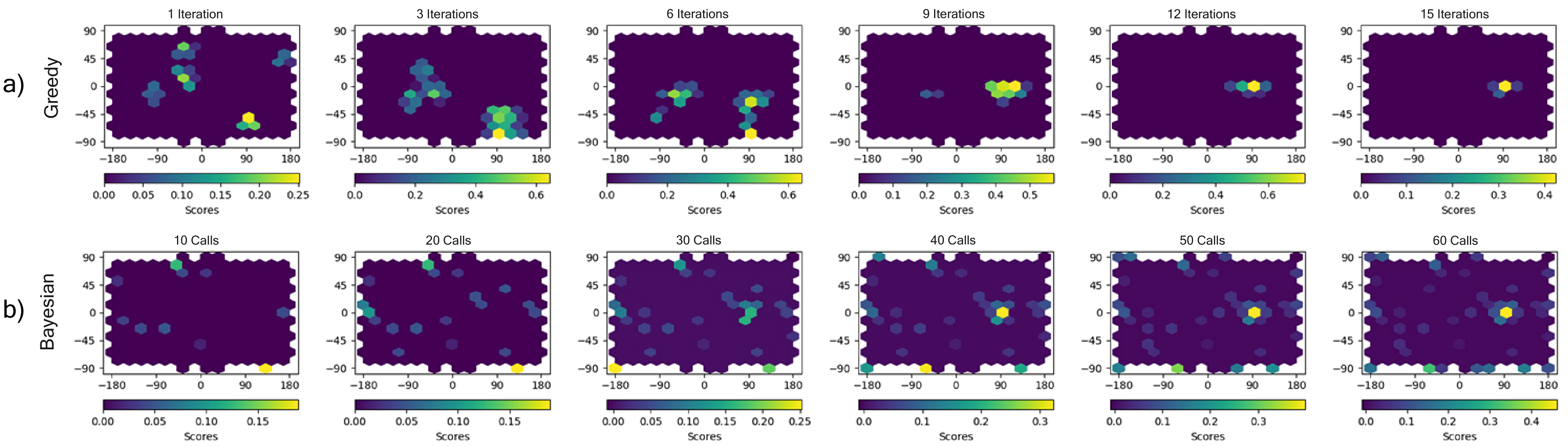}
         \caption{A single run of the search for the respective search algorithms a) greedy, b) Bayesian, on a randomly selected car object from the ShapeNet data set~\cite{chang2015shapenet} given the search query \textit{a picture of a car from the left}.}
      \label{fig:search_trajectory_analysis}
     \end{subfigure}
     \caption{\textit{top:} Distribution on object-only queries, \textit{center:} retrieved optimal viewpoints on CLIP PRE-TR and RC-HNS, \textit{bottom:} Execution of search algorithms.}
\end{figure*}

\paragraph{Comparison of Score Distributions for Object Only Queries.}
 To understand which viewpoints CLIP scores best on an object-only query such as \textit{a picture of a car}, we compare these object-only queries for all object categories tested on respective 3D objects from the test set. This tells us which viewpoints CLIP associates most with a given object category. Figure~\ref{fig:pure_object_queries_comparison} indicates that a PRE-TRained CLIP model is not able to distinguish specific viewpoint queries from pure object queries. 


\paragraph{Comparison of Optimal Viewpoints.}
Figure~\ref{fig:clip_compare_screenshots_car} shows the viewpoint images obtained from the optima of the scoring distributions generated by a CLIP model and a CLIP-RC-HNS model. The images illustrate that descriptions of viewpoints are indeed a bias in CLIP. 


Figure~\ref{fig:optimal_viewpoint_images_overview} illustrates the viewpoints resulting from the global optima of the scoring functions obtained from the CLIP-RC-HNS model. 

\subsection{Search Algorithm Analysis}
\label{sec:search_algorithm_analysis_appendix}
In our work, we are particularly interested in the impact of the shape of the scoring function on the performance of various search algorithms. Section~\ref{sec:greedy_search_details} provides details on the implementation of greedy search. Section~\ref{sec:search_algorithm_behavior_on_sphere} illustrates how the search algorithms listed above perform their task on a sphere. 



\subsubsection{Greedy Search Implementation Details}
\label{sec:greedy_search_details}
We implement a greedy search algorithm as a representative for gradient-based approaches. 
The greedy search starts with a grid-based approach on the Goldberg polyhedron and always follows the region with the highest score. It tries to find the optimum by greedily selecting the highest scoring regions at each iteration and searching in their neighboring regions at the next iteration. 
The search is initialized with $k$ randomly selected starting points (here $k=6$) from the Goldberg polyhedron. 
In addition, a cutoff value $c$ must be chosen to determine how many grid points will be considered in the next iteration of the search. The cutoff value can be described as a relative percentage or as an absolute cutoff value.
After evaluating all viewpoints with respect to the given query, the next iteration is started by selecting the locations with the highest scores considering the selected cutoff. 
All obtained scores and their neighboring sample points from the Goldberg polyhedron are added to the list of investigated viewpoints.
After that, the next iteration is started. 
The neighborhood range $n$, which specifies the number of neighborhood grid points to be examined, can be adjusted.
The search can be terminated after $i$ iterations or when no new items have been added to the list of investigated viewpoints.
In summary, the greedy search is parameterized by: ($k,c,n,i$). We chose greedy search as a test algorithm for our benchmark to see how much gradient-based methods as candidate algorithms for the text-viewpoint retrieval task in a 3D environment depend on a smooth structure of the scoring function in their performance. We use a greedy nearest-neighbour heuristic, since the function is only defined at a fixed number of points due to the discretization of the search space. 

\subsubsection{Bayesian Search Implementation Details}
\label{sec:bayesian_search_details}
Bayesian optimization~\cite{mockus1994application} is used to estimate the optimum of a black-box function that is costly to evaluate. The algorithm updates its Bayesian prior based on the stepwise function values obtained, increasing the certainty that the regions are likely to be optima and therefore more likely to be explored than other regions of the black box function. Then, the number of samples from the regions of interest is increased accordingly. 
We construct the search problem as a Bayesian optimization as follows: The input of the search algorithm is a vector of size five describing the camera position on the hypersphere around the target object: $r,\theta ,\varphi, x, y $. In this parameterization, $\theta$ and $\varphi$ are spherical coordinates, $r$ is the distance to the center of the 3D object, and $x$ and $y$ are the orientations of the camera along the horizontal and vertical axes. The location of the optimum of the scoring function with respect to a query $\mathbf{q}$ depends on the rotation of the 3D object, which we only know is centered around $(0,0,0)$. Therefore, Bayesian search tries to find the optimum of the scoring function with respect to the properties of the 3D object at hand given the search query $\mathbf{q}$. 
For our benchmarks, we use the implementation of the Bayesian optimization algorithm in~\citet{https://doi.org/10.5281/zenodo.1157319}.

\subsubsection{Search Algorithm Behavior on Sphere}
\label{sec:search_algorithm_behavior_on_sphere}
The experiments in Section~\ref{sec:experiments} 
have shown that a smooth scoring function is advantageous for search algorithms in text-viewpoint retrieval. This section visually analyzes why this is the case by examining how the algorithms perform on a sphere around a target object. 
\begin{table}
\setlength{\tabcolsep}{3.1pt} 
\small
\begin{tabular}{l | ccc | cccc} 
 \toprule
 Model & \textbf{P@1} & \textbf{P@5} & \textbf{P@10} & \textbf{R@1} & \textbf{R@5} & \textbf{R@10} & \\ 
 \midrule
 PRE-TR & 0.111 & 0.056 & 0.050 & 0.017 & 0.040 & 0.066 & \\ 
 FT & 0.778 & 0.567 & 0.500 & 0.113 & 0.330 & 0.485 & \raisebox{-.05\normalbaselineskip}[0pt][0pt]{\rotatebox[origin=c]{90}{\textbf{car}}} \\ 
 RC-HNS & 0.944 & 0.778 & 0.644 & 0.136 & 0.432 & 0.592 & \\
  \midrule
 PRE-TR & 0.056 & 0.078 & 0.050 & 0.008 & 0.057 & 0.073 & \\ 
 FT & 0.778 & 0.500 & 0.433 & 0.112 & 0.297 & 0.424 & \raisebox{-.05\normalbaselineskip}[0pt][0pt]{\rotatebox[origin=c]{90}{\textbf{airpln}}} \\
 RC-HNS & 0.833 & 0.522 & 0.439 & 0.119 & 0.310 & 0.441 & \\
 \midrule
 PRE-TR & 0.000 & 0.045 & 0.033 & 0.000 & 0.030 & 0.042 & \\ 
 FT & 0.500 & 0.322 & 0.339 & 0.074 & 0.217 & 0.400 & \raisebox{-.05\normalbaselineskip}[0pt][0pt]{\rotatebox[origin=c]{90}{\textbf{mbike}}}\\
 RC-HNS & 0.667 & 0.500 & 0.450 & 0.098 & 0.312 & 0.462 & \\
 \midrule 
 PRE-TR & 0.056 & 0.033 & 0.017 & 0.008 & 0.024 & 0.024 & \\ 
 FT & 0.389 & 0.311 & 0.294 & 0.056 & 0.179 & 0.286 & \raisebox{-.05\normalbaselineskip}[0pt][0pt]{\rotatebox[origin=c]{90}{\textbf{mug}}}\\
 RC-HNS & 0.667 & 0.489 & 0.483 & 0.097 & 0.312 & 0.532 & \\
 \midrule
 PRE-TR & 0.000 & 0.011 & 0.006 & 0.000 & 0.008 & 0.008 & \\ 
 FT & 0.667 & 0.511 & 0.439 & 0.097 & 0.312 & 0.465 & \raisebox{-.05\normalbaselineskip}[0pt][0pt]{\rotatebox[origin=c]{90}{\textbf{bench}}} \\
 RC-HNS & 0.944 & 0.744 & 0.689 & 0.136 & 0.411 & 0.592 & \\
 \bottomrule
\end{tabular}
\caption{Precision and recall metrics on synthetic data for the models \textit{PRE-TR, FT, RC-HNS} on the objects \textit{car, airplane, motorbike, mug, benchs} for\textit{front, back, left, right, top, bottom} viewpoints.} 
\label{tab:precision_and_recall_on_synthetic_data_category_granularity}
\end{table}

Figure~\ref{fig:search_trajectory_analysis} illustrates how the different algorithms approach the regions with higher scores differently. 
The greedy search with a low cutoff spreads across the sphere in waves, starting from the initial points. Once it touches a high point, it remains attached to it. In this respect, a good initialization is important, e.g., through a high number of random starting points. 
Bayesian search also starts from randomly initialized starting points around the hypersphere. Compared to greedy search, it reaches the optimum much faster and more purposefully, since sampling is not bound to any local constraints, such as neighboring regions. Another advantage over greedy search is that random starting points have much lower cost than in greedy search, since they do not cause additional computations in the following iteration. The figure shows that the focus of sampling from random starting points across the sphere leads to small, concentrated regions with high scores. In terms of success rate, Bayesian search is less prone to confounding optima, since a certain number of samples are drawn randomly from different regions anyway. Therefore, the approach is more robust to cases with multiple optima, as is the case with the CLIP-FT model. Despite these obstacles, a solution is reached relatively quickly. However, if the scoring function has a ragged structure like the CLIP-PRETR model, even a sampling-based approach has difficulty identifying the optimal regions due to the raggedness and non-uniformity of the function. 


\begin{table}
\setlength{\tabcolsep}{2.9pt} 
\small
\begin{tabular}{l | ccc | cccc} 
 \toprule
 Model & \textbf{P@1} & \textbf{P@5} & \textbf{P@10} & \textbf{R@1} & \textbf{R@5} & \textbf{R@10} & \\ 
 \midrule
 PRE-TR & 0.500 & 0.500 & 0.467 & 0.025 & 0.125 & 0.233 & \\ 
 FT & 1.000 & 1.000 & 0.967 & 0.050 & 0.250 & 0.483 & \raisebox{-.05\normalbaselineskip}[0pt][0pt]{\rotatebox[origin=c]{90}{\textbf{car}}} \\
 RC-HNS & 1.000 & 0.933 & 0.950 & 0.050 & 0.233 & 0.475 & \\
  \midrule
 PRE-TR & 0.333 & 0.367 & 0.350 & 0.017 & 0.092 & 0.175 & \\ 
 FT & 1.000 & 1.000 & 0.917 & 0.050 & 0.250 & 0.458 & \raisebox{-.05\normalbaselineskip}[0pt][0pt]{\rotatebox[origin=c]{90}{\textbf{airpln}}} \\
 RC-HNS & 1.000 & 0.833 & 0.750 & 0.050 & 0.208 & 0.375 & \\
 \midrule
 PRE-TR & 0.167 & 0.300 & 0.300 & 0.008 & 0.075 & 0.150 & \\ 
 FT & 0.667 & 0.633 & 0.650 & 0.033 & 0.159 & 0.325 & \raisebox{-.05\normalbaselineskip}[0pt][0pt]{\rotatebox[origin=c]{90}{\textbf{mbike}}}\\
 RC-HNS & 0.833 & 0.733 & 0.783 & 0.0417 & 0.183 & 0.392 & \\
 \midrule 
 PRE-TR & 0.167 & 0.167 & 0.167 & 0.008 & 0.042 & 0.08 & \\ 
 FT & 1.000 & 1.000 & 0.967 & 0.050 & 0.250 & 0.483 & \raisebox{-.05\normalbaselineskip}[0pt][0pt]{\rotatebox[origin=c]{90}{\textbf{mug}}}\\
 RC-HNS & 0.833 & 0.933 & 0.933 & 0.042 & 0.233 & 0.467 & \\
 \midrule
 PRE-TR & 0.333 & 0.200 & 0.167 & 0.0167 & 0.050 & 0.083 & \\ 
 FT & 1.000 & 0.733 & 0.583 & 0.050 & 0.183 & 0.292 & \raisebox{-.05\normalbaselineskip}[0pt][0pt]{\rotatebox[origin=c]{90}{\textbf{bench}}} \\
 RC-HNS & 0.667 & 0.500 & 0.500 & 0.033 & 0.125 & 0.250 & \\
 \bottomrule
\end{tabular}
\caption{Precision and recall metrics on real data for the models \textit{PRE-TR, FT, RC-HNS} on the objects \textit{car, airplane, motorbike, mug, benchs} for\textit{front, back, left, right, top, bottom} viewpoints.} 
\label{tab:precision_and_recall_on_real_data_category_granularity}
\end{table}

\subsection{Retrieval Metrics Analysis}
\label{sec:retrieval_metrics_analysis}
Table~\ref{tab:precision_and_recall_on_synthetic_data_category_granularity} shows the precision and recall metrics on \textbf{synthetic data} broken down by object category. Table~\ref{tab:precision_and_recall_on_real_data_category_granularity} shows the precision and recall metrics on \textbf{real data} obtained from the LAION-5B data set.


\end{document}